\documentclass{article}
\usepackage{PRIMEarxiv}
\usepackage[utf8]{inputenc} 
\usepackage[T1]{fontenc}    
\usepackage{hyperref}       
\usepackage{url}            
\usepackage{booktabs}       
\usepackage{amsfonts}       
\usepackage{nicefrac}       
\usepackage{microtype}      
\usepackage{amsmath}        
\usepackage{hhline}         
\usepackage{adjustbox}      
\usepackage{algorithm}      
\usepackage{algpseudocode}
\usepackage{lipsum}
\usepackage{fancyhdr}       
\usepackage{graphicx}       
\usepackage{subcaption}
\usepackage{caption}
\usepackage{subcaption}
\usepackage{newtxtext,newtxmath} 
\usepackage{tabularx}
\usepackage{booktabs}
\usepackage{multirow}
\usepackage{array}
\usepackage{orcidlink}      
\graphicspath{{media/}{Images/}}
\usepackage{lipsum}
\usepackage{xcolor}
\usepackage{tikz}
\usetikzlibrary{shapes.geometric, arrows.meta, positioning}
\usepackage{wrapfig}
\usepackage{graphicx}

\title{Two-Valued Symmetric Circulant Matrices: Applications in Deep Learning}

\author{
  Jayakrishna Amathi \orcidlink{0009-0008-5064-5663} \\
  Department of Computer Science and Engineering\\
  University of North Texas\\
  Denton, TX, USA.\\
  \texttt{amathijayakrishna@gmail.com} \\
   \And
  Venkata Prasanth Yanambaka \orcidlink{0000-0003-4625-8050} \\
  Division of Computer Science\\
  Texas Woman's University \\
  Denton, TX, USA.\\
  \texttt{vyanambaka@twu.edu} \\
  \And
  Saraju P. Mohanty \orcidlink{0000-0003-2959-6541} \\
  Department of Computer Science and Engineering\\
  University of North Texas\\
  Denton, TX, USA.\\
  \texttt{smohanty@ieee.org} \\
  \And
  Elias Kougianos \orcidlink{0000-0002-1616-7628} \\
  Department of Computer Science and Engineering\\
  University of North Texas\\
  Denton, TX, USA.\\
  \texttt{elias.kougianos@unt.edu} \\
}

\begin{document}
\maketitle

\begin{abstract}

Despite the success of deep neural networks in vision, medical diagnosis, and IoT scenarios, their deployment on resource-limited platforms poses serious challenges due to their high storage requirements, computational complexity, and large footprint. In particular, fully connected layers require a large number of weights, making it difficult for edge devices to accommodate them. To overcome these challenges associated with limited platforms, this paper proposes the Two-Valued Symmetric Circulant Matrix (TVSCM), a very sparse architecture that employs just two weights per layer to keep it circulant and symmetric. The extreme form of structured sparse architecture provides negligible storage costs compared to traditional full-weight storage. Instead of hardware and additional stages of other traditional sparse learning techniques, such as low-rank approximation and pruning approaches, this architecture provides an extreme form of sparsity, achieving very minimal storage requirements. The simulation study demonstrates more than 80$\times$ reduction in model parameters, reducing parameters from 623,290 to 7,852 on MNIST and from 24,709 to 942 on the MIT-BIH arrhythmia dataset, while maintaining comparable accuracy from 97.6\% to 93.5\% on MNIST and from 97.6\% to 93.1\% on MIT-BIH. Due to its minimal architectural requirements and very low power consumption, this architecture would be ideal for edge computing platforms, tiny-ML platforms, IoMT systems, and battery-powered systems.

\end{abstract}


\section{Introduction}

The healthcare domain is currently undergoing a rapid transformation to intelligent systems. The massive amount of health information generated per unit time is driven by billions of wearable sensors and IoT health devices. Health IoTs, including wearables such as smartwatches and implant biosensors, are used to monitor health parameters, identify irregularities, and assist clinicians in early disease diagnosis \cite{sundaravadivel2017everything, greco2020trends}. But most of these health monitors rely on the cloud for processing. In life-critical applications, a small delay in decision-making can be unsafe \cite{amin2020edge}.

To overcome these challenges, scientists have recently begun to migrate computation from remote servers in the cloud to Edge/Fog devices, which are more proximate to the areas where the data is being generated. This development has become known as \textit{Edge Intelligence}\cite{sundaravadivel2017everything,greco2020trends}. It has enabled smart sensors, smartphones, and local gateways to analyze data locally, thereby delivering health responses both faster and more accurately. For example, wearable ECG and blood sugar level health monitors can analyze their data independently, without relying on remote networks\cite{greco2020trends}.

Nevertheless, applying complex deep learning architectures on small-scale edge systems remains a challenge. This is because deep neural networks have millions of parameters, which are memory- and compute-intensive, unaffordable for small-scale medical equipment \cite{sundaravadivel2017everything}. Hence, a similar emphasis has been given to designing light, parameter-efficient AI models. The use of structured mathematical representations, such as circulant and symmetric matrices, has recently shown immense potential for reducing the complexity of neural networks \cite{rancea2024edge}. Thus, the direct implementation of intelligence in small-scale Edge Systems, such as those used in hospitals, homes, and health-tracking devices, becomes possible\cite{hayyolalam2021edge}.

As such, this study proposes a deep learning architecture based on the Two-Valued Symmetric Circulant Matrix (TVSCM) for use in the healthcare domain. The mathematical structure of the proposed model is based on the presence of exactly two symmetrical values in a circulant configuration. The feasibility of the proposed Two-Valued Symmetric Circulant Matrix (TVSCM) is validated on the MNIST and MIT-BIH Arrhythmia databases \cite{mitbih_arrhythmia_2005}, achieving an 80-fold reduction in parameter count compared to the conventional model while maintaining the same level of accuracy.

This paper is organized as follows: Section \ref{SEC:Overview} presents an overview of matrices. Section \ref{SEC:DeepLearning} provides the background of deep learning, Section \ref{SEC:DeepLearning_in_Healthcare} discusses the role of deep learning in healthcare, and Section \ref{SEC:Challenges_of_Deep_Learning} highlights the challenges of deep learning in healthcare. The research across multiple disciplines in similar areas is discussed in Section \ref{SEC:Related_Research}. Section \ref{SEC:Contributions_of_Current_Paper} outlines the contributions of the current paper. Section \ref{SEC:TVSCM} presents the proposed Two-Valued Symmetric Circulant Matrix (TVSCM) model. Section \ref{SEC:Results} presents the experimental results. Finally, Section \ref{SEC:Conclusion} concludes the paper and discusses future research directions.
\begin{figure}[ht]
\centering
\begin{tikzpicture}[
    node distance=1.3cm and 1.6cm,
    every node/.style={font=\small},
    ellipse_node/.style={ellipse, draw, thick, align=center, minimum width=2.0cm, minimum height=1.1cm},
    rect_node/.style={rectangle, draw, thick, align=center, minimum width=2.2cm, minimum height=1.1cm},
    arrow/.style={-{Stealth[length=3mm]}, thick}
]

\node[ellipse_node] (wearable1) {Wearable\\Sensors};
\node[rect_node,  right=of wearable1] (cloud) {Cloud\\Processing};
\node[ellipse_node, right=of cloud] (decision1) {Health\\Decision};

\draw[arrow] (wearable1) -- (cloud);
\draw[arrow] (cloud) -- (decision1);

\node[below=0.3cm of cloud, font=\small] (label1) {(a) Traditional cloud-dependent approach (slow, high latency)};

\node[ellipse_node, below=2.2cm of wearable1] (wearable2) {Wearable\\Sensors};
\node[rect_node,   right=of wearable2] (tvscm) {TVSCM\\Edge Model};
\node[ellipse_node, right=of tvscm] (decision2) {Health\\Decision};

\draw[arrow] (wearable2) -- (tvscm);
\draw[arrow] (tvscm) -- (decision2);

\node[below=0.3cm of tvscm, font=\small] (label2) {(b) Proposed TVSCM-based Edge Intelligence approach (fast, low latency)};

\end{tikzpicture}
\caption{Healthcare inference: traditional cloud-based vs.\ proposed TVSCM edge-based approach.}
\label{fig:intro_diagram}
\end{figure}

\section{Overview of Matrices}
\label{SEC:Overview}

\begin{figure}[h!]
    \centering

    \begin{subfigure}[b]{0.28\textwidth}
        \centering
        \includegraphics[width=\textwidth]{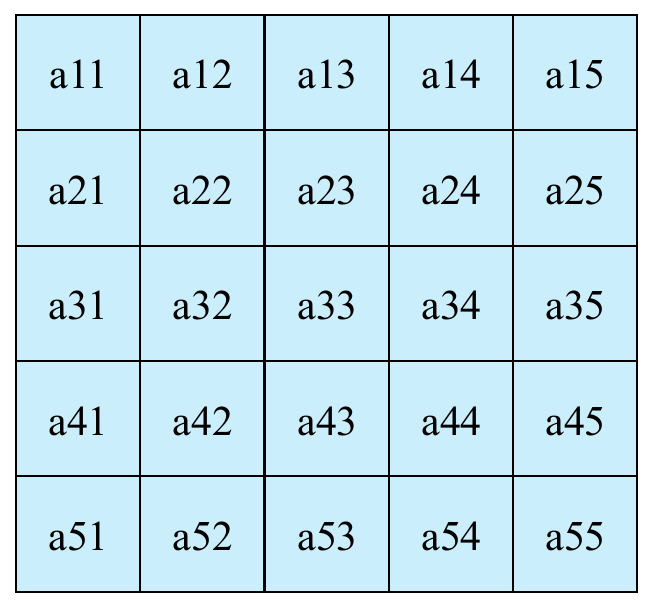}
        \caption{\textbf{Dense Matrix}}
        \label{fig:dense_matrix}
    \end{subfigure}
    \hfill
    \begin{subfigure}[b]{0.28\textwidth}
        \centering
        \includegraphics[width=\textwidth]{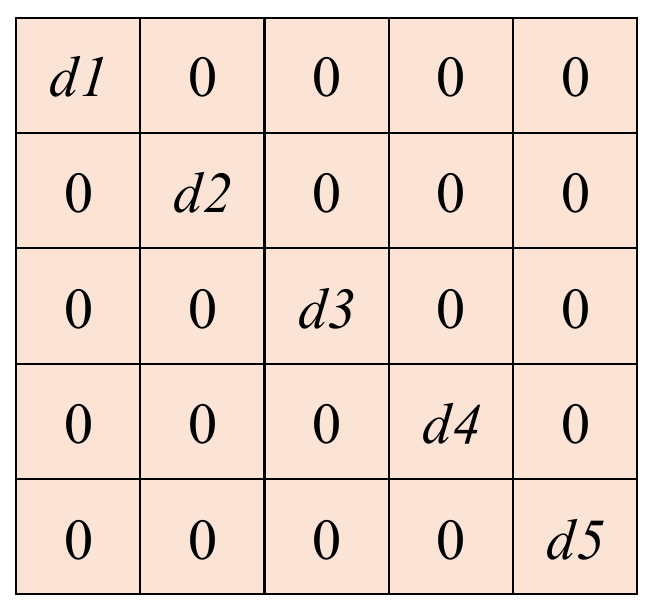}
        \caption{\textbf{Diagonal Matrix}}
        \label{fig:circulant_matrix}
    \end{subfigure}
    \hfill
    \begin{subfigure}[b]{0.28\textwidth}
        \centering
        \includegraphics[width=\textwidth]{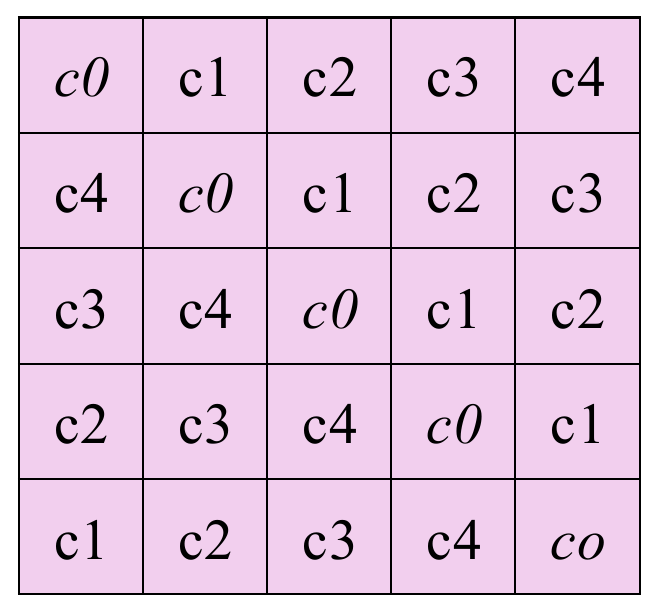}
        \caption{\textbf{Circulant Matrix (rows shifted)}}
        \label{fig:diagonal_matrix}
    \end{subfigure}

    \vspace{0.5cm}

    \begin{subfigure}[b]{0.28\textwidth}
        \centering
        \includegraphics[width=\textwidth]{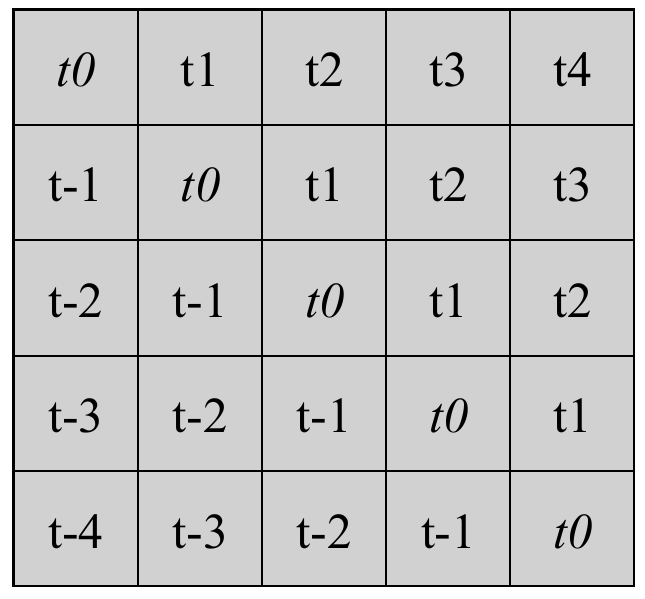}
        \caption{\textbf{Toeplitz Matrix (constant diagonals)}}
        \label{fig:toeplitz_matrix}
    \end{subfigure}
    \hfill
    \begin{subfigure}[b]{0.28\textwidth}
        \centering
        \includegraphics[width=\textwidth]{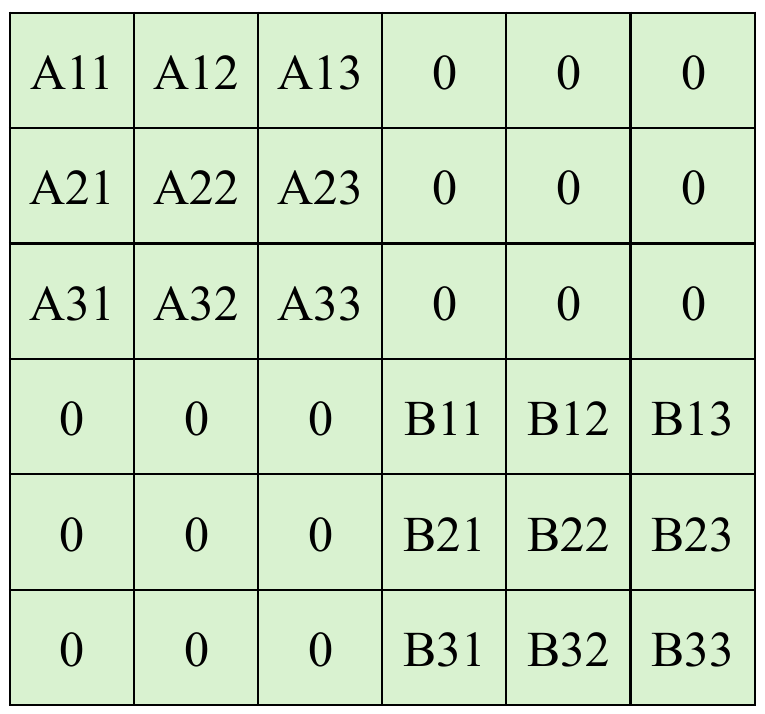}
        \caption{\textbf{Block-Structured Matrix (outlined blocks)}}
        \label{fig:block_structured_matrix}
    \end{subfigure}

    \caption{\textbf{Types of structured matrices highlighting parameter reduction.}}
    \label{fig:structured_matrices_overview}
\end{figure}

Matrices are the foundation of modern-day computational systems and mathematical modeling. They contain numerical data in an organized format amenable for transformation, linear transformation, and minimization of functionals in numerous applications of science. Fig. \ref{fig:structured_matrices_overview} depicts an exhaustive description of notable classifications of matrices and the relationships between them, while Table \ref{tab:Matrix_Types} describes crucial features, such as sparsity, symmetry, and time complexity, of each type of matrix. Understanding these categories of matrices is crucial for the design of efficient algorithms of deep learning, signal processing, and numerical computation \cite{serre2010matrices}. 

\begin{table}[h!]
\caption{Parameter comparison across matrix structures for general $n$.}
\label{tab:Matrix_Types}
\centering
\renewcommand{\arraystretch}{1.4}
\setlength{\tabcolsep}{8pt}
\begin{tabular}{|c|c|c|c|}
\hline
\textbf{Matrix Type} & \textbf{Params (n×n)} & \textbf{Key Property} & \textbf{Use in Machine Learning} \\ 
\hhline{|=|=|=|=|}
Dense & $n^{2}$ & Full Flexibility & Fully Connected Layers \\ \hline
Diagonal & $n$ & Independent scaling & Normalisation \\ \hline
Toeplitz & $2n-1$ & Const. Along diagonals & Convolutions \\ \hline
Circulant & $n$ & Cyclic Shifts & Compression, FFT \\ \hline
Block-Structured & varies & Partioned Structure & Efficient Compression \\ \hline
\textbf{Two-Valued Symmetric Circulant} & \textbf{2} & \textbf{Symmetric, Commutative} & \textbf{Proposed Model} \\ \hline
\end{tabular}
\end{table}

\subsection{Dense Matrices}

Dense matrices are matrices for which most of the elements are nonzero. They form the foundation of linear algebra, and they are the basis of the matrix, inversion, and decomposition operations. Although dense representations are fully expressive, they come along with substantial computation and memory overheads, costing typically $O(n^{3})$ operations for inversion or decomposition \cite{bioli2025preconditioned}. Dense matrices are often replaced by structured or sparse matrices by scientific computing and deep neural networks to boost scalability. They are, nonetheless, irreplaceable for full-rank problems and low-dimensional systems where exact arithmetic should be performed.

Bartels--Stewart method and the Krylov subspace approaches are the solver methods specifically designed for dense coefficient matrices present in systems of Sylvester or Lyapunov equations \cite{bioli2025preconditioned}. These strategies, although offering significant numerical precision, become impractical for large problems and, consequently, induce the discovery of the structured variants explained later.

Apart from the classical dense solvers, recent research has focused on improving the scalability of dense matrix computations while exploiting the underlying structure in the matrices. Several research papers show that a large number of dense matrices used in scientific computations and integral equation methods exhibit low-rank structure in their off-diagonal blocks. Dense solvers can be developed to minimize memory requirements and computational cost while maintaining stability. Randomized and matrix-free methods show promising results in solving large-scale dense matrices. These methods are efficient for solving large, dense matrices because they eliminate the need for explicit matrix construction. They only rely on matrix-vector product operations. Dense matrix solvers can be developed to reduce the memory requirement while maintaining stability\cite{liu2016parallel}. 

The second direction in the development of efficient dense matrix solvers is the application of hierarchical methods or fast multipole methods. These methods can be used either as direct solvers or preconditioners. Dense matrices can be converted into a sparse matrix representation. Near-field interactions can be handled separately, while the far-field interactions can be represented in a low-rank matrix. The application of fast multipole methods can reduce the time complexity of the solvers from cubic to linear or quasilinear for a wide range of dense matrices. When these methods are used as preconditioners, the convergence of the iterative solvers can be accelerated while maintaining accuracy\cite{coulier2017inverse}.

\subsection{Diagonal Matrices}

Diagonal matrices are the most basic matrices, defined by the property of having zero elements in off-diagonal locations. Due to this structure, they have simple computation of the eigenvalues and determinants, making diagonal matrices highly useful for analytical results and fast numerical algorithms \cite{serre2010matrices}. For machine learning, diagonal settings are applied to assist processes like parameter initialization, preconditioning, and covariance modeling, for which variable independence assumption plays an essential role.

Significant practical application regards permuted diagonal matrices, maintaining computational efficiency while, at the same time, providing flexibility by the rearrangement of the elements they contain. If considered along with compactly efficient architectures, the matrices significantly reduce both the storage and the multiplication costs while maintaining the steadiness of the whole deep network layers \cite{bioli2025preconditioned}. Diagonal matrices' combination of simplicity and flexibility make them key elements of both the theoretical and the practical aspects.

Furthermore, the ease of use of diagonal matrices has significant implications for the interface between theoretically efficient and practically used data structures, such as those used in parallel computations. A considerable body of literature has been devoted to the feasibility of transforming large block matrices into diagonal form, thereby reducing computational complexity. This topic is also pertinent within the context of an implicit numerical scheme, as diagonalization provides an opportunity for the replacement of computationally expensive block inversions by a series of scalar computations\cite{pulliam1981diagonal}.

Significant investigations in the field have established the viability of diagonal and permuted-diagonal matrices as a foundational basis for efficient algorithms compatible with hardware architectures. The capacity of diagonal-dominant matrices to optimize matrix elements through permutations or factorization provides a means of optimizing cache utilization and parallel processing. Hence, not only are diagonal matrix data structures beneficial for traditional numerical methods, but they are also relevant for data-centric applications and can provide a foundation for lightweight data structures with predictable memory access patterns\cite{rossi1999parallel}.

\subsection{Toeplitz Matrices}

Toeplitz matrices have constant diagonals, and each diagonal from left to right of the descending type has the same value. They intrinsically arise in time-series analysis, digital signal processing, and control systems \cite{gray2006toeplitz}. Classical theorems like Szegö's theorem give their asymptotic distributions of the eigenvalues and contain profound information on the system's stability and filtering. These matrices give us computational advantages because often the multiplication and inversion could be carried out even in $O(n^{2})$ or even $O(n \log^{2} n)$ time through algorithms based on the Fast Fourier Transform (FFT) \cite{ye2016every}.

Applications go beyond differential equations, covariance modeling, and stationary stochastic processes, where the Toeplitz forms are correlations of lag difference dependencies \cite{gray2006toeplitz}. Strong connections between Toeplitz and circulant matrices enable their fast approximation through FFTs, making them popular components of the current signal and image analysis pipelines.

The circulant and Toeplitz matrices' efficiency mainly originates in the compatibility of these matrices with the Fast Fourier Transform (FFT), transforming convolution actions into elementwise products. Computational complexity decreases by transforming it from $O(n^{2})$ to $O(n \log n)$, and numerical stability remains \cite{ye2016every}. In the core review by Gray, the observation of circulant matrices, as they are diagonalizable by the discrete Fourier transform, facilitating highly efficient filtering processes and eigenvalue computations, is presented \cite{gray2006toeplitz}. These FFT-based formulations constitute the link between the Toeplitz and the circulant structures, enabling compactness and reduced arithmetic, respectively. These benefits realized by FFT actions directly trigger the shift toward circulant and block-circulant constructions for the large-scale neural systems, where the fast spectral convolution remains indispensable for scaling.

\subsection{Circulant Matrices}

Circulant matrices are the special type of Toeplitz matrices where each subsequent row is the cyclic shift of the current or the preceding. They have the characteristic property $C = \text{circ}(c_{0}, c_{1}, \ldots, c_{n-1})$, which enables diagonalization by the Discrete Fourier Transform (DFT), transforming complex matrix multiplications into elementwise products \cite{gray2006toeplitz}. It is this property upon which their use for convolutional operations, spectral measurements, and neural network compression is founded.

Circulant matrices are also asymptotically equivalent to Toeplitz matrices, meaning their eigenvalue distributions converge as the matrix size increases \cite{gray2006toeplitz}. These matrices provide extreme efficiency in both storage—requiring only $O(n)$ parameters—and computation—$O(n \log n)$ using FFTs. Their structured simplicity, as shown in Fig. \ref{fig:circulant_matrix}, and predictable algebraic properties make them central in recent model compression frameworks, where block-circulant weight matrices have reduced network parameters by several magnitudes while maintaining accuracy \cite{gray2006toeplitz}. 

Circulant and block-circulant matrices have been utilized as efficacious replacements for unstructured dense weight matrices in deep learning. CircCNN demonstrates applying block-circulant weights to enable storage reduction from $O(n^{2})$ to $O(n)$ and computation complexity reduction from $O(n^{2})$ to $O(n \log n)$ via FFT-based multiplication \cite{ding2017circnn}. This structured scheme abolishes the requirement of retraining upon compression and accomplishes $5\times$--$9\times$ acceleration along with negligible accuracy loss across models including LeNet-5 and AlexNet \cite{ding2017circnn}. Correspondingly, PermDNN model applies permuted diagonal and circulant structures for guaranteeing regular connectivity and energy-efficient computation upon inference \cite{deng2018permdnn}. These findings confirm the idea that the theoretical benefits of structured algebra are transferrable to hardware-friendly neural compression schemes.

\subsection{Block-Structured Matrices}
\label{sec:Block_Structured_Matrices}

Block-structured matrices have data organized as blocks or submatrices, and heterogeneous blocks of dense, diagonal, Toeplitz, or circulant components are permitted. Block structuring gives the flexibility of modeling complex systems like multi-layered neural networks or multi-dimensional PDE discretizations \cite{bioli2025preconditioned}. Block-circulant structures are particularly balanced between expressiveness and structure, and they preserve the FFT-based computational benefits within matrix partitions.

Block-circulant weight matrices have been successfully utilized in deep learning to make model sizes smaller and inference faster. Block Toeplitz and block diagonal structures are also found in covariance estimation of big data and in numerical solutions of structured linear systems \cite{gray2006toeplitz}. These structural factorizations retain low memory demands and enable parallel computation, as summarized in Table. \ref{tab:Matrix_Types}. 

\subsubsection{Block-Circulant Architectures in Neural Networks}

Block-circulant architectures divide neural weight matrices into the smaller circulant sub-blocks, allowing for a fine-grained trade-off between accuracy and compression\cite{ding2017circnn}. Each of the submatrices ensures FFT compatibility, so both convolutional and fully connected layers exploit the same underlying spectral operation\cite{Gorbett2024}.CircCNN theoretical analysis further demonstrates that, asymptotically, block-circulant neural networks have the same representational capacity as uncompressed networks\cite{ding2017circnn}. This design ensures an uncompromising mathematical basis for compression ratios, doing away with heuristic pruning strategies. Block partitioning, as demonstrated by the Fig.~\ref{fig:block_structured_matrix} of the Circnn paper, achieves linear scalability of hardware design while retaining model expressiveness, making these architectures the foundation for fast and efficient deep network design.

\subsubsection{Parallelization and Hardware Efficiency}

At the hardware level, block-circulant structures are highly valued for their ability to enhance memory locality and parallelism significantly. CirCNN architecture combines FFT as the base computing block, using recursive decomposition to enable universal and small-footprint implementations throughout the convolutional and fully connected layers\cite{ding2017circnn}. FPGA and ASIC implementations illustrate that whole deep networks like AlexNet are packable into a few megabytes of on-chip memory because of block-circulant compression, reducing high-energy accesses to the DRAM\cite{ding2017circnn}. Likewise, PermDNN framework achieves $4\times$ reduction of arithmetic computations and $3.9\times$ energy efficiency improvement by exploiting block-permuted diagonal matrices along with optimized processing elements\cite{deng2018permdnn}. These results, together, highlight the way structured matrix forms map directly into hardware acceleration benefits measurable and quantifiable.

\subsection{Symmetric and Low-Rank Matrices}
Symmetric matrices, for which $A = A^{T}$, ensure real eigenvalues and orthogonal eigenvectors, crucial characteristics for stability and optimization. They appear in covariance matrices, energy functionals, and kernel methods of scientific computation. The preconditioned low-rank Riemannian optimization methodology broadens applications by targeting symmetric positive definite (SPD) systems, and the computation becomes tractable by using low-rank manifold optimization\cite{bioli2025preconditioned}.

Low-rank matrices approximate large systems by keeping the largest modes or components only. This drastically reduces the storage and computation, and it becomes feasible to obtain the solution where the full matrices are too large to be stored in memory\cite{bioli2025preconditioned}. These approaches are the backbone of the vast majority of modern deep learning and scientific computing algorithms, where approximate factorization or Kronecker-based preconditioners have significant efficiency gain. Symmetry and low-rank structure, combined, ensure the scalability and robustness needed of modern systems.
\subsubsection{Symmetric Matrices and Eigen Decomposition}
Symmetric matrices, as well as the specific case of symmetric positive definite systems, play an essential role in optimization as well as the stability of large-scale computations. In the context of the proposed Preconditioned Low-Rank Riemannian Optimization algorithm, the use of SPD matrices ensures the definition of inner products as well as the promotion of convergence for manifold-based optimization solvers\cite{bioli2025preconditioned}. The symmetry of the proposed matrices ensures the existence of real eigenvalues as well as orthonormality. These are essential factors for the stability of the proposed optimization algorithms\cite{bioli2025preconditioned}. In the context of the specific application of deep model compression, the stability of the optimization process is as important as the simplification of the associated computational complexity.
\subsubsection{Low-Rank Approximations for Model Compression}
The use of low-rank matrices helps achieve efficient representation of high-dimensional data by emphasizing dominant features in the data structure, thus resulting in significant savings in storage and computation. The placement of this problem within the setting of Riemannian optimization, as a maximization (or minimization) problem of a quadratic form over a set of specific rank manifolds, enables flexibility in rank selection for the purpose of efficiency\cite{bioli2025preconditioned}. Accordingly, for the Efficient Computation exploration, rank-based decomposition by singular value trimming demonstrated keeping only the principal singular components, maintaining accuracy while significantly improving runtime\cite{wang2025efficient}. These results confirm low-rank modeling achieves the perfect balance between accuracy and compression, and it serves the theory behind the deep learning architectures of fewer parameters.

\section{Deep Learning Background}
\label{SEC:DeepLearning}
For better understanding of the suggested approach, in this part, the basic terms of deep learning and the architectural principles of neural networks are explained. Also, some approaches used for minimizing the number of parameters to make the model applicable are mentioned.
\subsection{Neural Network Architectures}
\begin{figure}[h!]
    \centering
    \includegraphics[width=0.7\textwidth]{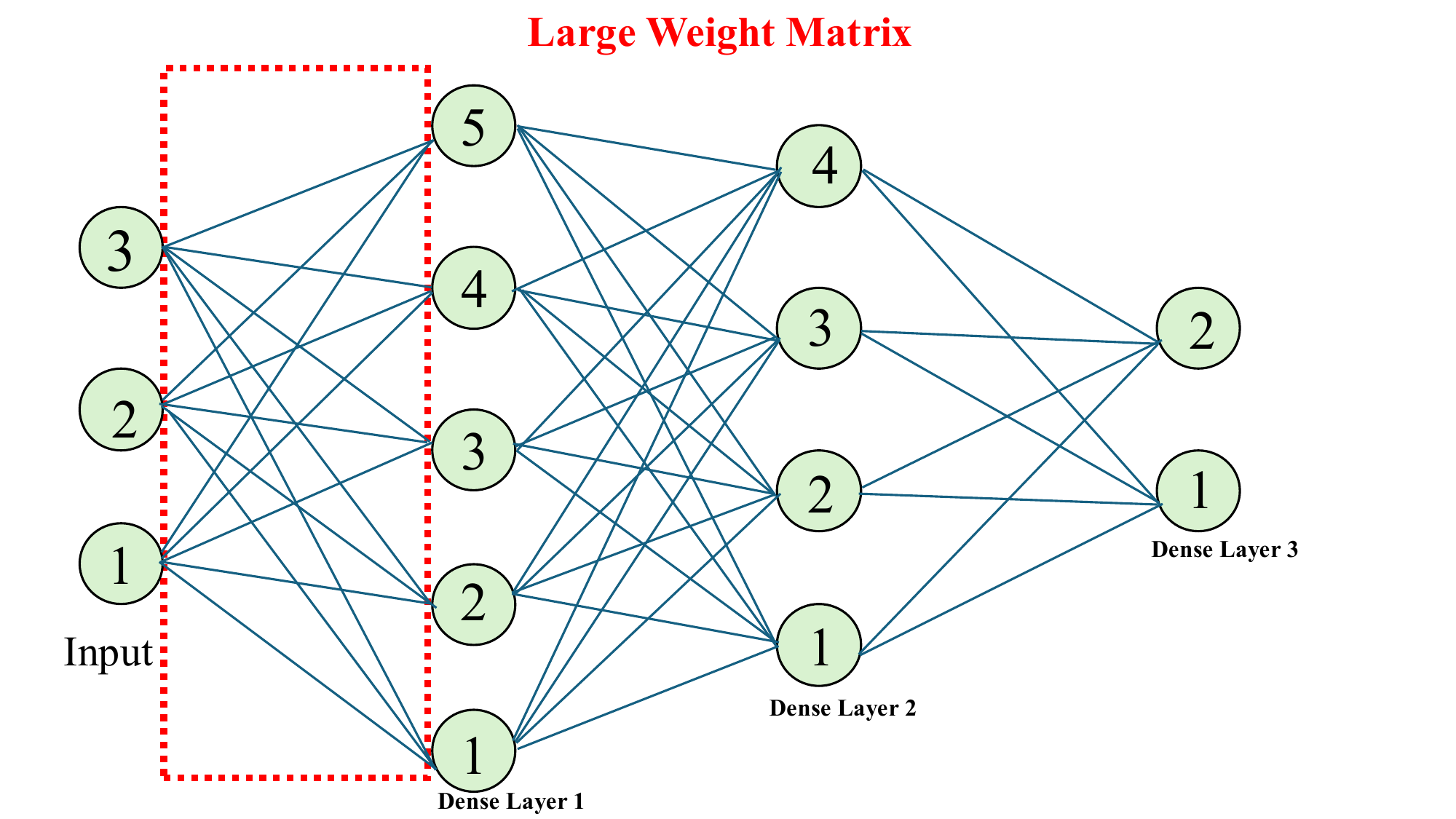}
    \caption{Typical deep neural network with parameter-heavy dense layers.}
    \label{fig:Deep_Neural_Network}
\end{figure}

Neural networks consist of layers of interconnected neurons that are capable of transforming input data. These neurons use weighted computations and nonlinear activation functions to change the input data. The typical network consists of an input layer where data are input, hidden layers where abstract features are learned, and an output layer where the network makes the prediction. Networks become capable of learning complex patterns by increased depth and width but are prone to increased parameter explosion, energy dissipation, and overfitting\cite{spasov2019parameter, dalavai2024improving}.This process is illustrated in Fig.~\ref{fig:Deep_Neural_Network}, which shows a typical deep neural network architecture.

Optimization of architectures demonstrated that well-crafted layer structures lower parameter numbers significantly. Separable and grouped convolutions yield strong predictive ability and low redundancy\cite{spasov2019parameter}. Multi-task learning architectures enhance the generality of models, particularly for small or heterogeneous datasets\cite{spasov2019parameter}. For the tasks of classifying images, the preprocessing of normalization, noise elimination, and histogram equalization of data accelerates convergence and representation of features,leading to stronger accuracy using fewer resources \cite{dalavai2024improving}.

\begin{figure}[h!]
    \centering
    \includegraphics[width=0.63\textwidth]{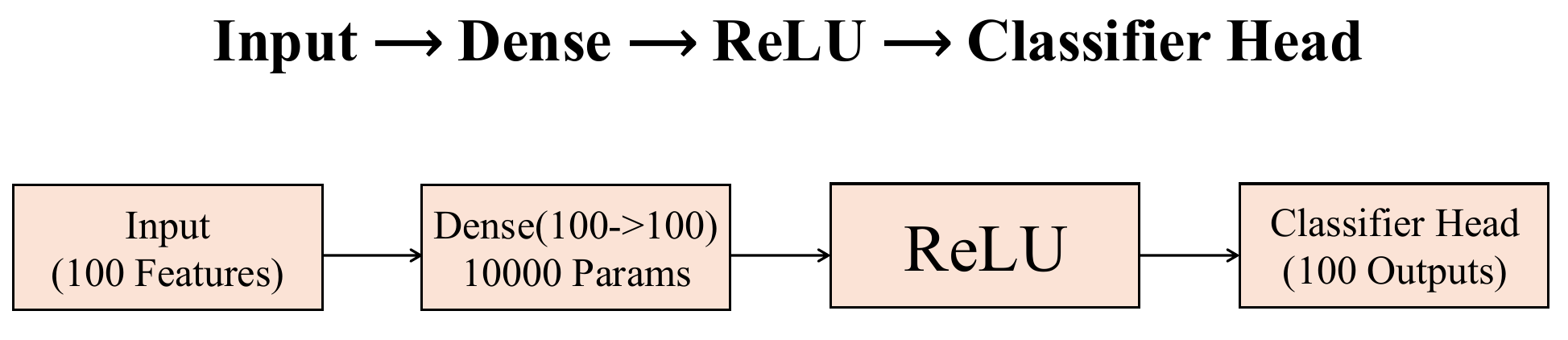}
    \caption{Typical neural network architecture with fully connected (dense) layers.}
    \label{fig:Neural Network Architecture}
\end{figure}

Autoencoder-based designs have increased efficiency by reducing high-dimensional data to compact latent representations and producing outputs with minimal loss reconstruction. For the purpose of modeling distributed physical systems, these architectures have replaced expensive numerical solvers by learning reduced-order mappings of system states and outputs\cite{franco2023deep, wang2016deep}. These developments demonstrate how the design of lightweight architectures—by exploiting separable convolutions, feature augmentation, and latent-space compression—is reducing network complexity while maintaining or even improving accuracy. The overall neural network structure, consisting of an input layer, a dense layer with a ReLU activation, and a classifier head, is illustrated in Fig.~\ref{fig:Neural Network Architecture}.

\subsection{Parameter Reduction in Deep Learning}
Parameter-reduction methods supplement architectural efficiency by reducing the number or accuracy of learnable weights directly. They decrease storage and computation requirements, allowing the application of these models to embedded and handheld systems without severe predictive loss.

\subsubsection{Pruning}
Pruning means the removal of unnecessary or lower-order parameters, and it achieves sparse representations which are compact and have lower latency. Unstructured pruning aims to remove certain connections individually, and structured pruning aims to remove full filters or neurons, which aligns better with hardware constraints. If pruning is followed by fine-tuning, the accuracy remains the same as the baseline model. Besides, the combination of pruning and any other compression methods further achieves the best balance between model size and model quality\cite{cheng2023survey,marino2023deep}.

Recent research has also shown that pruning involves not only removing weaker weights from the already-trained model. A significant number of research studies have also shown that the architecture of the pruned network plays an important role in the final performance. In many cases, training the pruned network from scratch has yielded better performance than fine-tuning the large pre-trained model. This has also given insight into the fact that pruning might be treated as an alternative to designing the network architecture. Further research has also shown the importance of pruning across the dataset, training, and evaluation\cite{blalock2020state,liu2018rethinking}.

\subsubsection{Quantization}
Quantization reduces numerical precision by representing the weights and the activations with lower-bit numbers, compared to the standard 32-bit floating-point format\cite{coquelin2024optimization}. Quantization, as a technique, cuts down the memory required significantly and computation costs but maintains equivalent accuracy by means of proper calibration methods. Quantization-based models run with high efficiency on dedicated processors and on the edge, with low power usage; therefore, they are applicable in real-time inference scenarios where resources are limited\cite{marino2023deep}.

Existing literature has conceptualised quantization as a well-regulated process by which continuous or high-precision values are converted to a specific set of values with minimal error. The basic analysis of the quantization theory has established the trade-off between the two variables; however, the trade-off has been well-managed through a very detailed and calibrated approach. Recent analysis of the machine learning domain has established the potential of neural networks to perform well with lower numerical values, and the quantization has been well-managed through the distribution and sensitivity of the layers. The analysis has also established the potential of lower-bit quantization to perform well in terms of computation and power, making it more viable\cite{gray2002quantization,berezin1974quantization}.

\subsubsection{Low-Rank Factorization}
Low-rank factorization approximates large weight matrices using the product of smaller matrices, thereby cutting parameter counts and floating-point operations. Techniques that learn optimal ranks for each layer during training achieve significant compression with minimal loss of accuracy\cite{idelbayev2020low}. Incorporating rank regularisation during training, instead of post-optimization, enhances performance stability and eliminates the necessity for retraining. These low-rank methods produce compact and fast networks suited for mobile and edge deployment\cite{xu2019trained}.

Following the premise presented in the above sections, several research papers prove the efficiency of the low-rank factorization approach in improving the efficiency of the training process. Hence, the training algorithm avoids the repetitive process of matrix factorization, thereby operating on smaller matrices, which in turn reduces the overall computational time. Moreover, the existing literature shows that learning the rank during training can help avoid over-compression, thereby increasing the model’s accuracy. Low-rank factorization has been proven to significantly reduce the number of parameters in large-scale neural networks, especially those with a high-dimensional output space\cite{sainath2013low,wen2012solving}.

\subsubsection{Structure-Preserving Methods}
Structure-preserving methods promote low-rank structures at training time without altering the topology of the network. Through the composition of linear layers and weight-decay regularization, the compactness of the weight matrices is learned by the model naturally. Post-training singular-value truncation results in final compression. It obviates pre-training or selection of the rank by-hand while ensuring architectural consistency and excellent accuracy across the dataset\cite{zhang2024structure}.

In accordance with this view, existing and recent research has highlighted that it is crucial to maintain the fundamental architecture of a system or model to ensure stability within it. Results from structure-preserving modelling techniques have shown that a system’s inherent structure can facilitate learning that maintains the constraints within that system, hence providing stability to the results. Rather than significantly changing a system’s fundamental structure, such models show promise in helping a system learn representations within its existing structure, hence providing stability to the results\cite{amathi2025note,bergen2007structure}.

\subsubsection{Domain-Specific Model Reduction}
For the engineering and scientific communities, the principal of parameter reduction underscores the importance of system representation simplifications and not the reduction of weights by itself. Autoencoder-based reduced-order models are capable of accurately capturing the principal dynamics of physically high-dimensional systems by exploiting low-dimensional latent spaces. These models are capable of reproducing system outputs successfully and significantly reducing the computation time, and consequently, enable simulation and control of complex distributed processes virtually in real time\cite{franco2023deep,wang2016deep}.

Building on this perspective, recent research shows that learning compact latent representations is essential for handling complex systems with many variables. Empirical research in reduced-order modelling suggests that autoencoders can maintain key system behaviours while removing redundant information, enabling faster simulations and more efficient control strategies. The literature in the field of representation learning and domain generalization also suggests that the removal of redundant or domain-specific information allows the model to focus more on the key system behaviours, thus allowing for better system performance. The above research clearly suggests that reducing system representations is, in fact, a viable solution for large-scale applications in engineering and science\cite{ding2022domain,zhang2023domain}.

\subsubsection{Quantum-Based and Hybrid Models}
\begin{figure}[h!]
    \centering
    \includegraphics[width=0.7\textwidth]{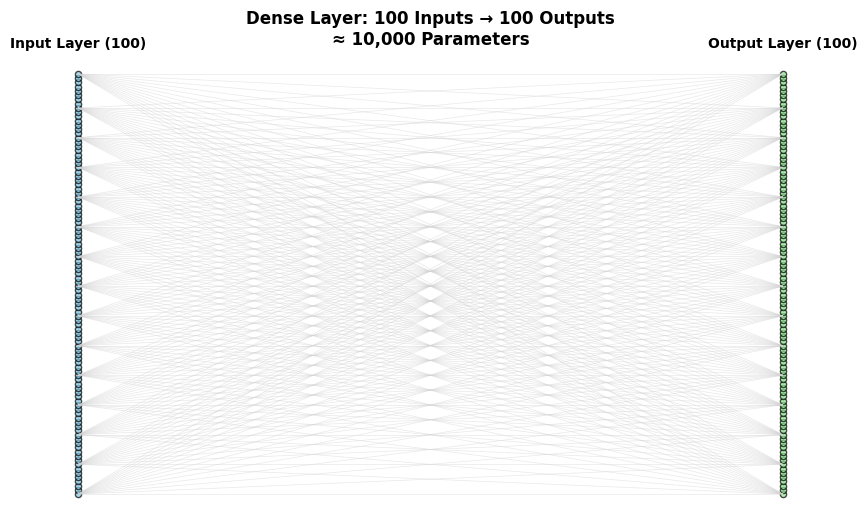}
    \caption{Dense architecture figure}
    \label{fig:Dense_Architecture_Figure}
\end{figure}

Quantum-inspired algorithms integrate the principles of quantum computing and deep learning to promote computation efficiency. Quantum-inspired neural networks utilize superposition and entanglement of quantum computation to achieve optimization and feature mapping through fewer operations. Experimental results demonstrate the ability of the hybrid algorithms to speed up training and enhance scalability, suggesting prospects of future parameter-efficient neural networks as the maturity of quantum hardware arrives. The parameter explosion occurring in dense layers, where every input connects to every output, is depicted in Fig.~\ref{fig:Dense_Architecture_Figure}.\cite{gonaygunta2024quantum}

\section{Deep Learning in Healthcare}
\label{SEC:DeepLearning_in_Healthcare}

Deep learning in recent years has emerged as one of the strongest technologies in medicine. It enables physicians and scientists to manage large-scale medical information like visual images, lab work, and gene information. In contrast with classical machine learning, deep learning can automatically capture effective patterns from original information without the requirement for the laborious choice of features. Its capability allows it to perform very well in disease diagnosis, in identifying risks in patients, and in formulating personalized medicine\cite{kothinti2024deep}.

\begin{figure}[h!]
    \centering
    \includegraphics[width=0.9\textwidth]{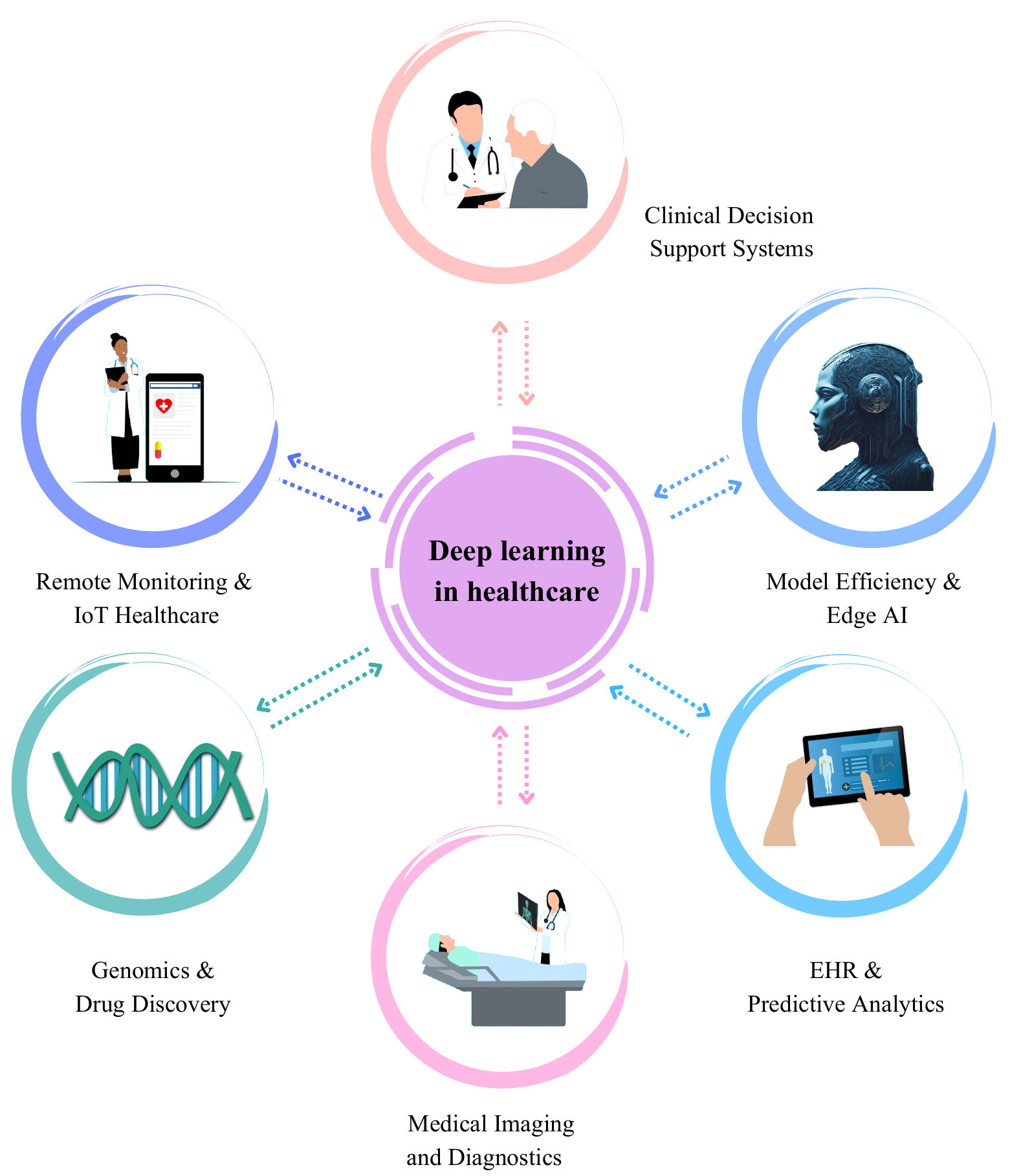}
    \caption{Deep Learning in healthcare.}
    \label{fig:Deep Learning in healthcare}
\end{figure}

The addition of deep learning transformed the work in healthcare organizations. Studies confirm that the neural networks can detect cancers, heart disease, and other illnesses earlier and more reliably than before. From reading medical scans and electronic health records to examining DNA and monitoring patients with online-connected devices, its use ranges widely. It has been observed by various researchers that the increasing trend of using high-performance computing and large healthcare data bases helps models operate faster and more accurately, thus assisting in more precise and informed decision-making by doctors and medical professionals. The overall relationship among the major application areas of deep learning in healthcare is shown in Fig.~\ref{fig:Deep Learning in healthcare} \cite{nazir2025deep,rahman2024machine}.

\subsection{Clinical AI for Diagnosis, Prognosis, and Decision Support}
Deep learning is believed to play a vital part in the field of medicine, particularly in the interpretation of medical images, including X-rays, magnetic resonance imaging, computed tomography scans, and microscopic images\cite{dash2022deep}.Deep learning models, including convolutional neural networks and transformers, are used in the detection of tumors, early disease prediction, and assisting medical professionals in the decision-making process. These models use general data patterns and operate as efficiently as human radiologists\cite{rahman2024machine}.

Deep learning is not only used for image analysis but also for improving electronic health records. Patient history, physiological signals, and laboratory results can be used to forecast future health conditions and the likelihood of hospital readmission using history-based models. LSTM networks and RNNs can be used for forecasting future health conditions, including heart failure and sepsis. Technologies like these assist healthcare professionals with early warnings and recommendations for certain interventions\cite{panahi2025deep}.

Deep genomics is the study of DNA and proteins, and the information they contain, in the hope of identifying mutations that cause illness and aiding the development of new medicines. This is a new application of deep learning, and it has the potential to recognize patterns that might otherwise be missed, thus aiding the development of new targets for drugs and cures\cite{kothinti2024deep}.

As a further extension of the applications of deep learning, current research is able to show that deep learning has the potential to combine all forms of information available in the world of medicine, from imaging and clinical information to biological information, and thus allow a more informed decision-making process to take place, allowing a more precise decision to be made by the physician regarding the patient\cite{magrabi2019artificial}.

\subsection{Connected and Real-Time Care (IoT and Wearable Applications)}
Deep learning plays an essential role in the development of advanced health care systems. Devices like hospital sensors, smartwatches, and health trackers are used to monitor patients’ health by obtaining real-time data like glucose levels, heart rate, blood oxygen saturation levels, among others. Long Short-Term Memory (LSTM) networks and Convolutional Neural Networks (CNNs) are used for processing the obtained data in real-time. They are used for detecting anomalies in the data. This has helped improve telemedicine services, which are beneficial for patients with chronic health conditions\cite{rahman2024machine}.

The next emergent domain is the amalgamation of deep learning and the Internet of Things (IoT) in healthcare. Here, networks consisting of smart sensors and edge devices collect medical information, conduct local processing, and push only valuable information to cloud systems. In this setup, the processing becomes faster and reduces the need for persistent internet connectivity. The technology finds usage in cardiac monitoring, diabetic applications, and in systems used for elder care\cite{rahman2024machine}.

Generative deep learning networks such as GANs and diffusion models are also utilized in creating synthetic medical information for the purpose of training AI systems. This protects patient confidentiality and addresses the issue of limited data availability for rare diseases\cite{kothinti2024deep}.

The present research aims to examine the potential of deep learning in improving the interpretation of complex health data obtained from various wearable devices. Unlike previous systems that used a single modality of health data, recent deep learning-based systems are able to use multiple modalities of health data, i.e., heart rate, motion, skin temperature, and oxygen saturation, for a comprehensive assessment of a person's health. This has the potential to aid in the monitoring of patients without the need for frequent hospital visits\cite{dian2020wearables}.

Furthermore, recent research shows that deep learning can be combined with edge- and cloud-based IoT platforms to deliver fast, reliable healthcare services. The combination of deep learning with edge- and cloud-based IoT platforms can improve the processing of sensor data by running local neural networks on wearable or edge devices. The processed data can then be communicated to cloud systems for long-term analysis. Such an approach can be used to deliver quick and reliable health services to patients, which is now commonly implemented through modern wearable devices and smart health systems\cite{assaad2025developing}.

\section{Challenges of Deep Learning in Healthcare}
\label{SEC:Challenges_of_Deep_Learning}
Many studies show that deep learning strategies have widespread applications in the health sector in applications such as disease diagnosis, patient tracking, and radiograph examination. Despite generating good results in controlled research settings,the strategies face considerable challenges in clinical settings in hospitals. According to the literature, deep learning models require large amounts of data, computing power, and memory space. Therefore, the application becomes inconvenient in hospitals that have inadequate computational infrastructure\cite{udegbe2024role,wubineh2024exploring}.

Other studies suggest that hospitals in small urban areas or emerging areas also lack the appropriate infrastructure to have advanced systems in place. In addition, the connection between artificial intelligence systems and hospital databases becomes poor, and hence the smooth flow of information is halted. Therefore, while deep learning has so much potential, its incorporation into everyday healthcare procedures remains problematic\cite{sun2019mapping,alanazi2022using}.


\begin{figure}[h!]
    \centering
    \includegraphics[width=1.0\textwidth]{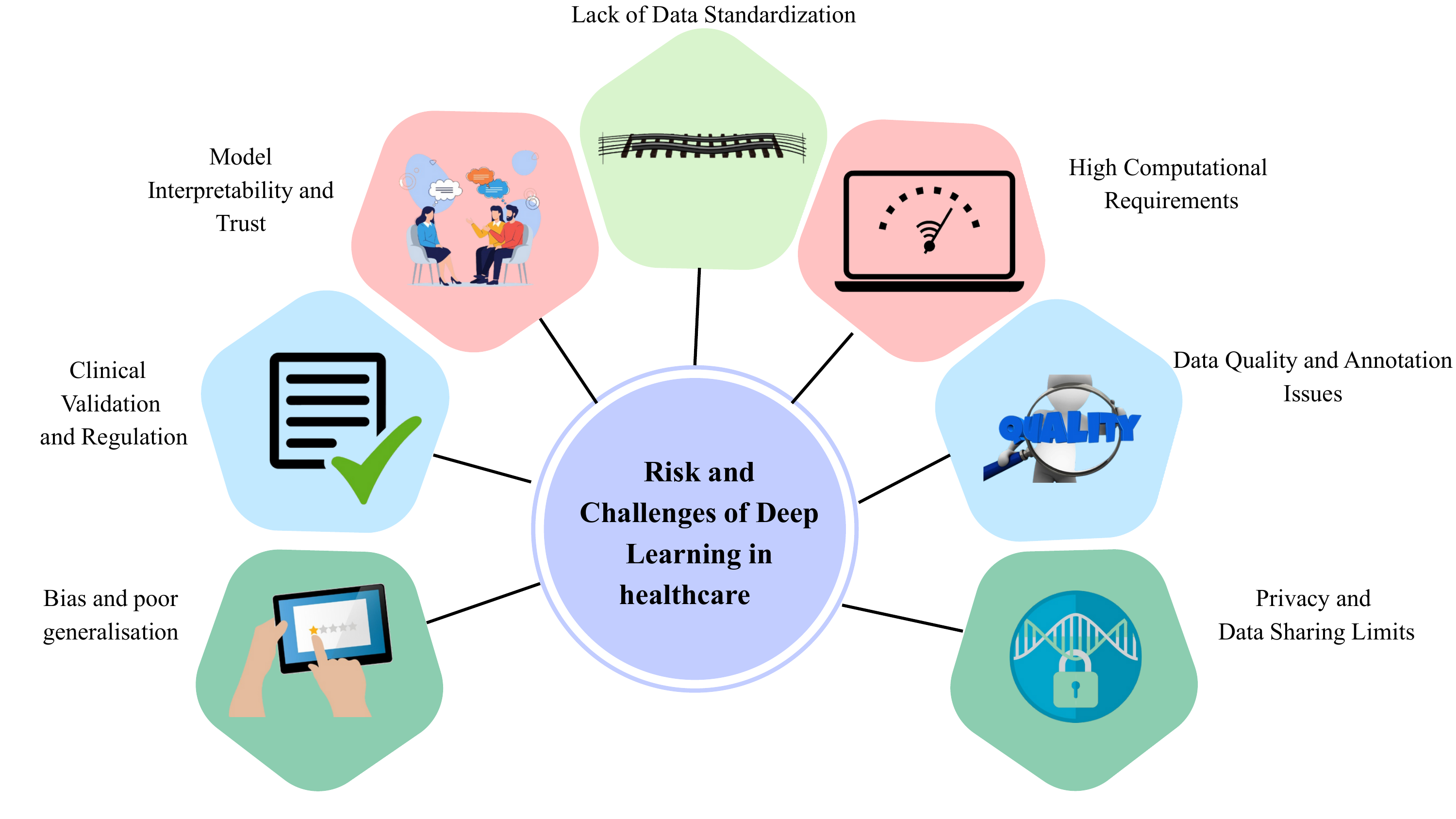}
    \caption{Challenges of Deep Learning in healthcare.}
    \label{FIG:methodology-diagram}
\end{figure}

\subsection{Computational and Infrastructural Challenges}
Deep learning models have millions of parameters. This means that they need enormous memory, exceptional computing power, and long training times. Most papers cite that training these models requires high-end hardware like GPUs or TPUs. Most hospitals which do not have such infrastructure could not afford to deploy these models with ease. This makes it hard to apply AI in real-time applications in healthcare where the results ought to be immediate\cite{alanazi2022using,amin2021healthcare}.

Additionally, several research works have proven that the use of large models on small devices such as smartwatches causes delays in information processing. This may be detrimental to patient care in emergency situations. Additionally, the inherent complexity of the models has often hindered physicians' inductive understanding of the underlying rationale for specific choices. Deep learning has thus posed several challenges to its implementation in resource-constrained environments in hospitals due to the large size and high cost of the models\cite{sun2019mapping,gerke2020ethical}.

\subsection{Data, Ethical, and Deployment Challenges}
Deep learning techniques need large amounts of data to be efficiently trained. However, data is not only sensitive but also located at various points in the health domain. In addition, empirical research suggests that data protection policies affect data sharing, which is a barrier to developing comprehensive data sets. In some cases, data is unbalanced, which can lead to inaccurate predictions for a given class of data. As such, these factors raise concerns about fairness and accuracy in artificial systems\cite{udegbe2024role,gerke2020ethical}.

The integration of hospital information systems with artificial intelligence systems is a factor that adds to the problem. Many hospital databases are out-of-date and have challenges in integrating with current AI technology. Such hindrances translate to lateness and inaccuracies in data communication. Also, several studies highlight the confusion revolving around where the liability lies in case of AI systems generating illegitimate outputs—by the developer, the healthcare facility, or the doctor. More issues include biases, lack of transparency, and a lower stakeholders' level of trust. All these data-related, ethical, and technical challenges make it hard to safely apply deep learning methods in real healthcare applications\cite{wubineh2024exploring,sun2019mapping}.
\subsection{Overview of Challenges}
The main challenges found in different studies are shown clearly in Fig.~\ref{FIG:methodology-diagram}. The figure presents an elementary illustration of the primary issues that are inherent with deep learning for healthcare. The figure indicates that problems associated with bias, poor generalization, lack of standardization, extensive computationally intensive requirements, minimal model interpretability, privacy limitations, and poor quality for data are all intertwined.
It also describes how the problems are correlated with each other through the figure. For example, if the data is not clean or balanced, then it causes model result bias. Likewise, if hardware is limited, then models cannot process vast medical data properly, leading to malfunction or slow processing. The two collectively present an easy method to comprehend the overall risk that hinders the proper and safe implementation of deep learning in hospitals\cite{dhar2023challenges}.These challenges are clearly summarized in Table.~\ref{tab:dl_challenges_noimpact}, which provides an overview of the major issues faced in deep learning for healthcare.

These problems mentioned above carry certain ramifications regarding the practical application of this approach. First, problems connected with interpretability and security may affect the confidence of the health-care specialist when it comes to decision making by the system. In addition to that, the considerable need for hardware, as well as some compatibility issues, might hinder the implementation of these systems in places that lack adequate technological and material resources. Hence, taking into account these problems is crucial for developing efficient deep learning solutions in the health-care sector.
\begin{table}[htbp]
\centering
\caption{{Summary of Key Challenges in Deep Learning for Healthcare}}
\begin{tabular}{|p{4cm}|p{12cm}|}
\hline
\textbf{Challenge Type} & \textbf{Description} \\ 
\hhline{|=|=|}
High Computational Demand & Deep learning systems need powerful computers, i.e., GPUs or TPUs, and have millions of parameters. Limited hospitals cannot easily deploy these systems\cite{alanazi2022using,amin2021healthcare}. \\ \hline
Memory and Power Limits & They are memory- and power-hungry. They are slow and inefficient to be executed on compact or wearable systems such as smart monitors or wearables\cite{udegbe2024role,sun2019mapping}. \\ \hline
Data Privacy and Security & Patient data sharing is limited because of data privacy laws. This limitation slows the amount of data available to build trustworthy models\cite{udegbe2024role,gerke2020ethical}. \\ \hline
Data Quality and Standardization & Medical data are typically incomplete, not cleansed properly, or stored in various formats. These make the data incorrect and decrease the accuracy\cite{alanazi2022using,amin2021healthcare}. \\ \hline
Model Interpretability & Numerous deep learning models function as "black boxes." Medical professionals find it challenging to comprehend the decision-making processes involved, resulting in diminished trust\cite{sun2019mapping,gerke2020ethical}. \\ \hline
Integration with Hospital Systems & Traditional hospital software and databases are poorly compatible with modern systems of AI. This drawback prevents smooth data interchange and real-time results\cite{udegbe2024role,wubineh2024exploring}. \\ \hline
Bias and Fairness Issues & When the training data isn't balanced, the machine learning models may discriminate between certain groups and provide biased outputs for the rest\cite{udegbe2024role,gerke2020ethical}. \\ \hline
Ethical and Legal Concerns & It is unclear who is responsible if AI makes a wrong decision. Also, there are not enough clear rules for safe and fair use\cite{wubineh2024exploring,sun2019mapping}. \\ \hline
\end{tabular}
\vspace{0.4em}
\label{tab:dl_challenges_noimpact}
\end{table}
\section{Related Prior Works}
\label{SEC:Related_Research}
The size and complexity of deep neural networks continue to increase. As a consequence, not only are memory requirements intensifying, but computation complexity and power requirements are increasing as well. In response to the challenging requirements of real-time and edge computing devices, there has been intense interest in efficient representations of matrix structures to maintain a low parameter count without sacrificing accuracy.

One common approach is to use circulant and block-circulant matrices as substitutes for full-weight matrices. The CirCNN model demonstrated that block circulants can reduce computational complexity and storage requirements from $\mathcal{O}(n^2)$ to $\mathcal{O}(n \log n)$ without sacrificing accuracy. However, this approach uses FFT computation to lower costs, but requires multiplications involving floating-point weights\cite{ding2017circnn}. The CircConv approach brought a circulant structure to convolution filters. The strategy reduced memory needs, but required full-precision multipliers\cite{liao2019circconv}. In accelerating a graph neural network model, block-circulants reduced computation and energy costs during compilation and execution on FPGAs but still relied on real weights and full-precision multiplications\cite{zhou2021blockgnn}. Real block-circulants in a Transformer model, paired with DCT-DST transforms, achieved up to a $41\%$ reduction in parameters but still relied on full-precision multiplications\cite{asriani2023real}.

Additional savings in parameters have also been analyzed in diagonal-circulant decompositions. The Deep Diagonal Circulant Neural Networks leverage alternating diagonal and circulant matrices to implement a compact model. However, this design still employs continuous parameters. The more extensive set of Low-Displacement-Rank (LDR) matrices, including Toeplitz-like transforms, proposed compressed models for a limited footprint in Machine Learning\cite{araujo2019understanding}. The computation still employs real-number arithmetic and cannot produce highly compressed versions required for compact AI\cite{sindhwani2015structured}.

Other methods utilize structured sparsity to alleviate storage and power consumption. In PERMDNN, permuted diagonal matrices were employed for simplicity in hardware computation and lower memory needs without resorting to FFT computation; nevertheless, circulant symmetry and minimum value parameters were not\cite{deng2018permdnn}. Parameter-efficient tuning has also adopted structured matrices. The Block-circulant adapters showcased extensive parameter reductions for fine-tuning large-scale models. However, changes were made only within adapter layers and not within entire architectures\cite{ding2025block}. Monarch matrices optimized training speeds via block-diagonal matrix representations. However, methods were still implemented using fixed-point arithmetic and focused primarily on training speeds rather than extensive architectures\cite{dao2022monarch}. SURM advanced and enhanced structured matrices for fine-tuning to promote high-quality performance with reduced parameters. Once more, changes were made in selective components. In all of these works, structured matrix design has been shown to reduce parameter count and improve computational efficiency for deep neural networks. However, the usage of real-valued parameters and multiplier computation is still widespread\cite{sehanobish2024structured}. 

One area that has not been extensively researched is weight representation with only two fixed constants while maintaining symmetry in circulant matrices to achieve extreme compression and reduced arithmetic computation for computing environments. In this context, this work is inspired by the goal of developing a compact architecture based on two-valued symmetric circulant matrices. This idea is highlighted in Table.~\ref{tab:structured_methods_comparison}, which compares prior structured-matrix approaches and positions the proposed model among them.

\begin{table}[ht]
\centering
\caption{Structured-Matrix Methods for Parameter Reduction in Deep Learning Models}
\label{tab:structured_methods_comparison}
\begin{tabular}{|p{1.5cm}|p{2.3cm}|p{2.3cm}|p{3.4cm}|p{2.6cm}|p{2.7cm}|}
\hline
\textbf{Prior Study} &
\textbf{Structured Representation} &
\textbf{Model Component} &
\textbf{Reduction Approach} &
\textbf{Efficiency Outcome} &
\textbf{Challenge} \\ 
\hhline{|=|=|=|=|=|=|}
CirCNN \cite{ding2017circnn} &
Block-circulant &
CNN / FC &
Combine all the blocks into a single vector and perform an FFT. &
Reduced $O(n^2)$ to $O(n\log n)$ &
Still depends on multipliers and floating-point numbers \\
\hline
CircConv \cite{liao2019circconv} &
Circulant kernels &
Conv layers &
Share Weights in Circulant Form &
Lower model size &
Calculations remain traditional multiplications \\
\hline
PERMDNN \cite{deng2018permdnn} &
Permuted diagonal &
Fully Connected &
Few parameters and no index overhead. &
High throughput &
Lack of circulant symmetry and compression capabilities. \\
\hline
DCNN \cite{araujo2019understanding} &
Diagonal + circulant &
Deep layers &
Factorize each dense matrix into two structured ones. &
Fewer parameters &
Sensitive to training and is still using real-valued weights. \\
\hline
BlockGNN \cite{zhou2021blockgnn} &
Block-circulant &
Graph neural network layers &
Compression by circulant blocks &
High speed and energy savings &
Requires support for FPGAs and real-valued weights. \\
\hline
LDR Transforms \cite{sindhwani2015structured} &
Low-displacement-rank &
Edge ML models &
Weight sharing through fast transforms &
3.5$\times$ compression &
Designed for moderate compression levels. \\
\hline
Real Block-Circ Trans \cite{asriani2023real} &
Real block-circ + DCT-DST &
Attention / FFN &
Use sparse weight matrices instead of full ones. &
Up to 41\% fewer parameters &
Multipliers are not completely eliminated; only partial compression is applied. \\
\hline
Monarch Matrices \cite{dao2022monarch} &
Block-diagonal product &
ViT / Transformers &
Structure for efficient hardware &
Faster training &
Focus on speed more than storage savings \\
\hline
Block-Circ Adapters \cite{ding2025block} &
Block-circulant adapters &
LLM fine-tuning &
Compress only adapter modules. &
14--16$\times$ fewer tuned params &
The base model still remains dense. \\
\hline
SURM \cite{sehanobish2024structured} &
Unrestricted LDR &
PEFT adapters &
Compact structured updates &
High accuracy with fewer params &
Compression is only on adapters. \\
\hline
\textbf{Proposed Model} &
\textbf{Two-valued symmetric circulant} &
\textbf{Full CNN / FC} &
\textbf{Binary-like weights + symmetry} &
\textbf{Ultra-high compression + no multipliers} &
\textbf{Broader evaluation needed} \\
\hline
\end{tabular}
\end{table}

Recent work in intelligent medicine has focused on developing smaller, faster neural network models suitable for deployment on portable or embedded medical devices. A case study on quantized convolutional neural networks for disease diagnosis from ultrasound images\cite{garifulla2021case} showed that model quantization reduced storage requirements by nearly $75\%$ and inference time by $97\%$, enabling real-time deployment on low-hardware-resource devices.

Saha et al. proposed a small one-dimensional CNN for ECG arrhythmia classification, achieving very high diagnostic accuracy while using much fewer trainable parameters compared to deeper models \cite{saha2025lightweight}. The smaller model showed good performance on databases such as MIT-BIH and PTB, making it suitable for use in embedded cardiac-monitoring devices.

Additional work on binary-weight convolutional neural networks\cite{akbari2018classification} also demonstrated that $32$-bit floating-point weights could be binarized, with around $32\times$ memory reduction at a loss of accuracy close to imperceptibility, which speaks volumes about the efficacy of binary or quantized forms for medical image analysis.

Therefore, these factors raise parameter reduction and storage minimization as essential factors for the efficiency of intelligent healthcare systems. Quantization, lightweight network design, and binary weight encoding facilitate the application of deep learning models on small, power-efficient medical devices. The proposed two-valued symmetric circulant model extends this line of work by providing a structured mathematical design that achieves even greater parameter reduction and computational efficiency, without sacrificing diagnostic accuracy. This comparison is clearly presented in Table.~\ref{tab:healthcare_comparison}, which shows how the proposed model differs from other efficiency techniques in smart-healthcare applications.

\begin{table}[htbp]
\small
\centering
\caption{Comparison of Model Efficiency Techniques in Smart-Healthcare Applications}
\label{tab:healthcare_comparison}
\renewcommand{\arraystretch}{1.1}
\begin{tabular}{|p{2.6cm}|p{2.8cm}|p{3.0cm}|p{2.8cm}|p{2.8cm}|}
\hline
\textbf{Prior Study} & \textbf{Technique / Model Type} & \textbf{Main Strategy for Compression} & \textbf{Reported Storage / Parameter Reduction} & \textbf{Remaining Challenge} \\
\hhline{|=|=|=|=|=|}
[1]Quantized CNNs\cite{garifulla2021case} & Quantized convolutional neural networks & Used lower-bit integer weights and activations rather than full-precision ones. & Approximately 75\% less storage and 97\% faster inference. & There is some inaccuracy for higher levels of quantization. \\ \hline
[2]Lightweight 1D CNN\cite{saha2025lightweight} & Small one-dimensional CNN for ECG data & Having fewer layers and filters without sacrificing accuracy. & Fewer parameters to estimate; better suited for embedded systems. & Not easily scalable for more complex signal inputs. \\ \hline
[3]Binarized-Weight CNNs\cite{akbari2018classification} & Binary-weight convolutional neural networks & Replaced 32-bit floating-point weights with $\pm 1$ binary values. & Approximately $32\times$ lower memory usage. & Possible accuracy drop in deeper models. \\ \hline
\textbf{Proposed Model} & Two-valued symmetric circulant matrices & Uses a structured matrix with only two constant weight values and symmetry & Extremely compact representation with minimal storage and no multipliers & Needs validation across multiple medical datasets \\ \hline
\end{tabular}
\end{table}

\section{Contributions of Current Paper}
\label{SEC:Contributions_of_Current_Paper}

Deep learning models transform healthcare operations by improving medical diagnosis methods, individual treatment strategies, and patient care decision systems\cite{kothinti2024deep}. But as these models grow larger, huge amounts of computation and memory are required making them difficult to fit in small devices and real-time systems. Both lightweight and accurate models are required to make AI more practical. Both lightweight and accurate models are required to make AI more practical\cite{wang2023e3d}. It is necessary to explore mathematical structures that drastically reduces parameters while preserving predictive power to meet this requirement \cite{ding2017circnn}\cite{kothinti2024deep}.

\subsection{Research Questions Addressed in the Current Paper}

Two valued symmetric circulant matrices(2VSCM) provide an opportunity to achieve this balance. The research questions addressed through this work are:

\begin{itemize}
    \item Using structured matrices, especially two-valued symmetric circulant matrices, can reduce the number of parameters in deep learning models while still keeping good accuracy.
    \item The unique mathematical properties in two-valued symmetric circulant matrices suit them well for usage in neural networks.
    \item In contrast with the prior high-density layers, the technique can maximize model efficiency, limit parameter usage, and maintain reasonable classification precision.
    \item This idea can be utilized in developing smaller and more comprehensible deep learning models which can perform well with bigger sets and in practice.
\end{itemize}
\subsection{Novel Contributions of the Current Paper}
\begin{figure}[h!]
    \centering

    \begin{subfigure}[b]{0.45\textwidth}
        \centering
        \includegraphics[width=\textwidth]{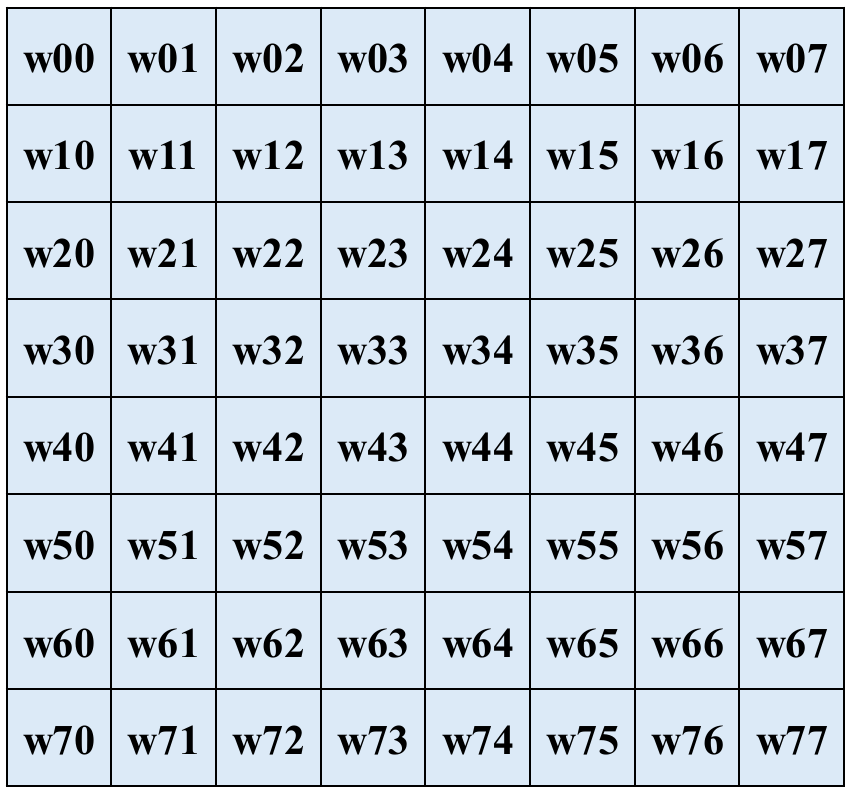}
        \caption{\textbf{Baseline Dense (Schematic)}}
        \label{fig:baseline_dense}
    \end{subfigure}
    \hfill
    \begin{subfigure}[b]{0.45\textwidth}
        \centering
        \includegraphics[width=\textwidth]{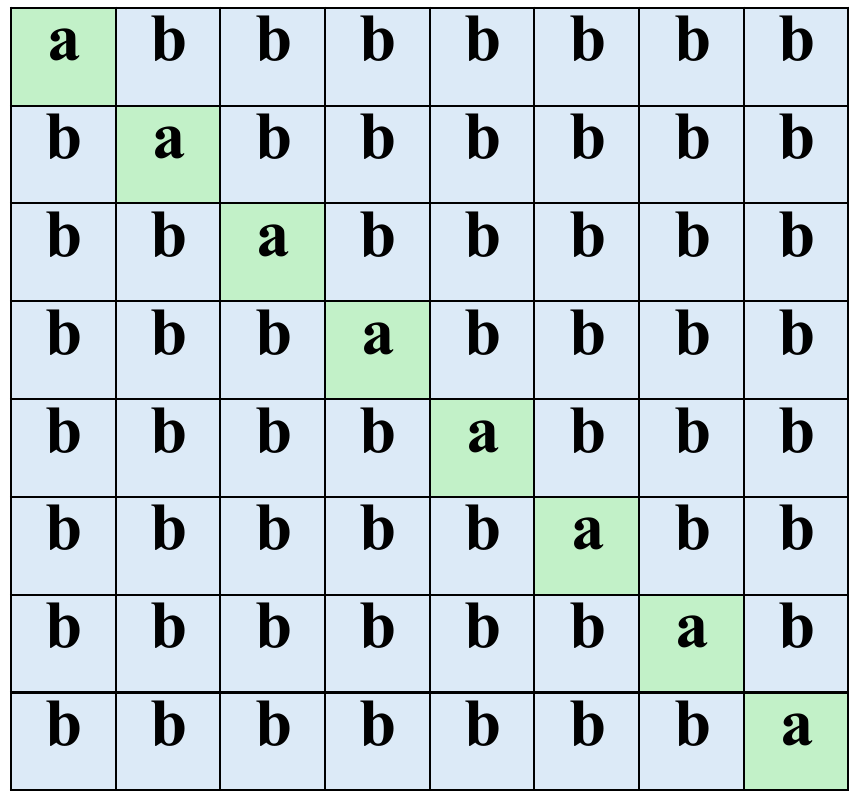}
        \caption{\textbf{Two-Parameter Circulant (Schematic)}}
        \label{fig:circulant_2param}
    \end{subfigure}

    \caption{\textbf{Comparison between a baseline dense weight matrix and the proposed two-parameter circulant matrix representation.}}
    \label{fig:comparison_baseline_circulant}
\end{figure}

Many existing models often reduce parameters but lose accuracy or are too complex. In this work, a two-valued symmetric circulant model is proposed to replace large dense layers with extremely compact structures. This idea is shown in Fig.~\ref{fig:comparison_baseline_circulant}, where the dense layer and the two-valued circulant structure are displayed side by side.
\begin{itemize}
    \item Introducing a new way to use two-valued symmetric circulant matrices to cut down the number of parameters in deep learning models.
    \item Providing a simple formula in mathematics to build a whole transformation matrix and explain its properties using just two values.
    \item The neural network architecture, which replaces dense layers with a circulant structure, aims to create a model that is smaller in size without compromising accuracy.
    \item Showing through experiments on larger datasets that the method decreases the parameters by approximately 80 times, with a slight decrease in accuracy.
    \item Laying the foundation for creating smaller, interpretable, and mathematically grounded models that can be applied to larger datasets and real-world problems.
\end{itemize}

\subsection{Significance of the Contribution}
The two-valued symmetric circulant method that is presented in this paper presents a new method that can be used to develop lightweight deep learning models without compromising their accuracy. The method is likely to be useful in the development of smart healthcare services, given their need to use wearable technology to ensure timely services. The method can be used to develop more effective services such as electrocardiogram (ECG) arrhythmia detection. The paper shows that two-valued symmetric circulant matrices can be used to develop effective AI models. It has laid the groundwork that can be used to conduct further research on the use of matrices to develop more effective AI models.

\section{Proposed Two Valued Symmetric Circulant matrices}
\label{SEC:TVSCM}

To address the drawbacks of conventional deep learning structures in resource-scarce scenarios, this section proposes the Two-Valued Symmetric Circulant Matrix (TVSCM) model. This model is based on the drawbacks of conventional structures that were discussed in earlier sections, such as the increase in parameters, memory, and the need for optimized computation on embedded systems as part of a deep learning (DL) stack. This model shows an improvement in feasibility when compared to traditional structures by its unique features, configuration, and computational characteristics. The following sections describe the development of TVSCM structures for Internet of Medical Things (IoMT).

\subsection{TVSCM-Based IoMT Architecture and Its Role in Healthcare Applications}

The existing healthcare infrastructure is evolving with the paradigm shift of the Internet of Medical Things (IoMT), which enables the integration of devices like wearable sensors, mobile health technologies, and cloud-based diagnostic solutions for continuous monitoring of the health condition of a patient. The overall architecture of IoMT environments is generally divided into two layers: a sensing layer, also known as a low-power physiological layer, and a cloud layer.
\begin{figure}[h!]
    \centering
    \includegraphics[width=0.9\textwidth]{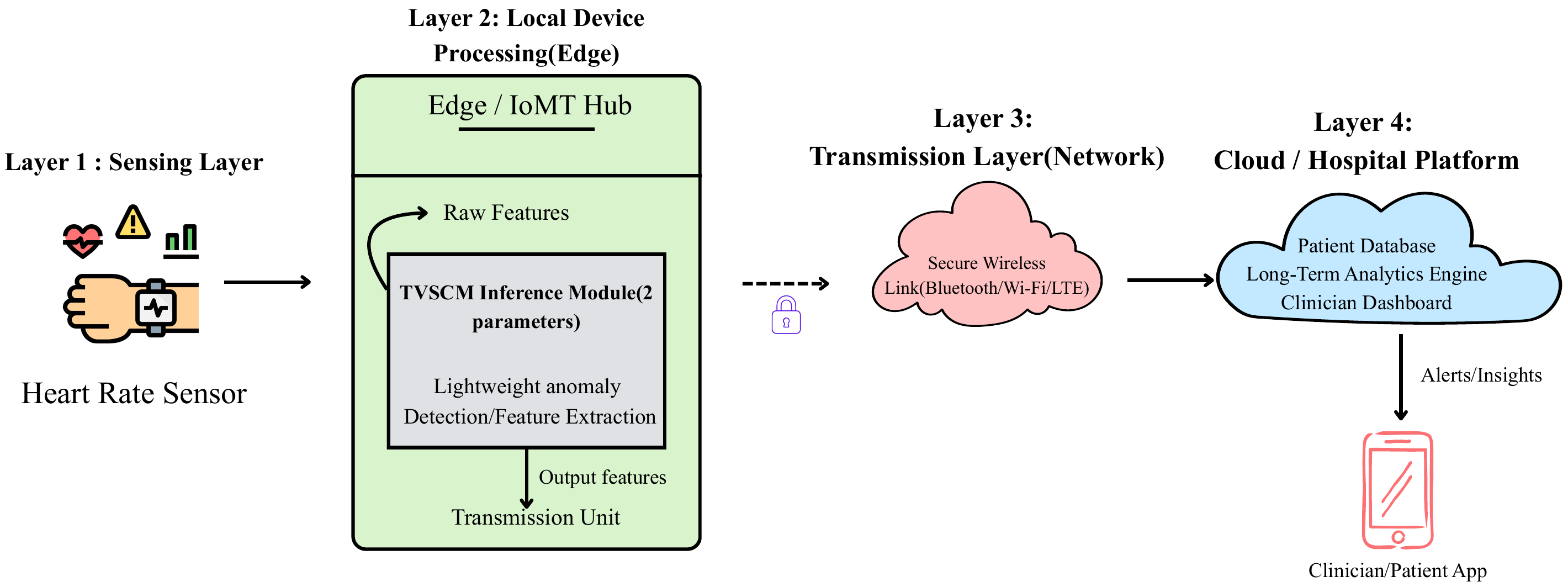}
    \caption{Proposed IoMT Architecture With TVSCM-Based Edge Intelligence.}
    \label{TVSCM_IOMT_Architecture}
\end{figure}
Despite the developments, IoMT environments face several issues, like the limited battery life and low memory capacity of wearable devices, which make it difficult to implement machine learning algorithms. Third, results generated by cloud computing can fluctuate intermittently due to connectivity constraints on one side and overall immediacy concerns on the other.The placement of the TVSCM module within the IoMT architecture is shown in Fig.~\ref{TVSCM_IOMT_Architecture}.

The above-mentioned challenges are directly overcome by the compressed CNN structure developed through TVSCM. Due to its minimal memory and computational overhead, TVSCM layers can be integrated into any IoT device deployed in the IoMT domain, such as wearables, patch monitors, and handheld diagnostic devices. The overall effect is to decrease the amount of communication required to transmit patient data to the cloud.

This architecture sees the raw physiological data, such as ECG, glucose, and activity, being extracted by the IoMT sensing layer. Rather than transmitting this raw data to the cloud for computing, the edge device uses TVSCM-based inference to perform lightweight classification/anomaly detection. Only insights/anomalies are transmitted to cloud computing for long-term analysis. This is a collective effect that yields timely results, increased dependability, enhanced privacy, and a significantly reduced energy budget.

\subsection{TVSCM Design Overview}

To achieve a compact and efficient representation of the weights in a neural network's layers, a new Two-Valued Symmetric Circulant Matrix (TVSCM) is proposed. In a standard dense layer, all elements of its weight matrices are learned independently, which means that a dense layer has a large number of learnable parameters. In a circulant matrix, a single row is enough to represent a whole matrix because all remaining elements in a matrix are cyclic shifts of its first row. Moreover, in a symmetric matrix, all elements are reflected across the main diagonal.

This compression is further accentuated when considering that all elements of the matrix are constrained to take only two different values, $a$ and $b$. This implies that the entire matrix is adequate to define a situation that needs only two variables and a rotation pattern. From a mathematical perspective, a symmetric circulant matrix, as described above, is fully defined by a few variables, as opposed to thousands of variables in a typical situation. This yields a significant reduction in the number of trainable parameters.

\begin{figure}[!t]
    \centering
    \includegraphics[width=0.35\textwidth]{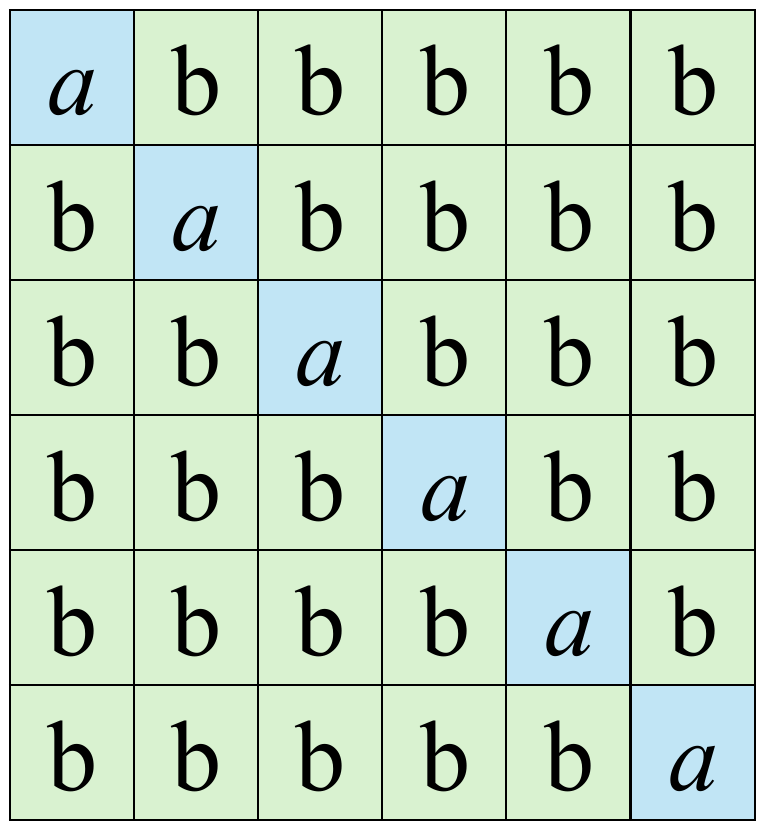}
    \centering
    \caption{Two-parameter symmetric circulant matrix structure.}
    \label{fig:TVSCM_Structure}
\end{figure}

Furthermore, circulant matrices can be used for the acceleration of the matrix multiplication process using the Fast Fourier Transform (FFT), which reduces the computational complexity of the process from quadratic to logarithmic.  This conjoint effect of structured symmetry, two-value constraints, and FFT-based computations makes TVSCM a preferential option for model compression and fast executions. This structure is shown in Fig.~\ref{fig:TVSCM_Structure}, which illustrates how the two-valued symmetric circulant matrix is arranged.
\subsection{Integration of TVSCM into Neural Network Pipeline}
In the proposed model, the traditional dense layers in a neural network can be systematically replaced with TVSCM layers, while all other components of the network remain unchanged. This would allow preserving the network's original learning structure and dynamics while minimizing the number of trainable parameters.

\begin{figure}[h!]
    \centering

    \begin{subfigure}[b]{0.45\textwidth}
        \centering
        \includegraphics[width=\textwidth]{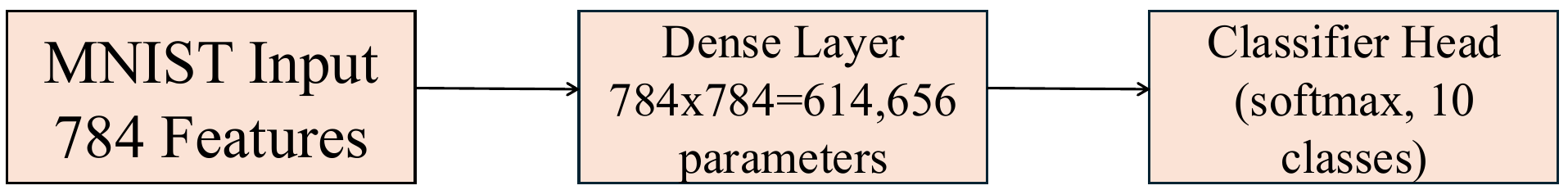}
        \caption{\textbf{Standard Dense Mapping (Parameter Explosion)}}
        \label{fig:standard_dense_mapping}
    \end{subfigure}
    \hfill
    \begin{subfigure}[b]{0.45\textwidth}
        \centering
        \includegraphics[width=\textwidth]{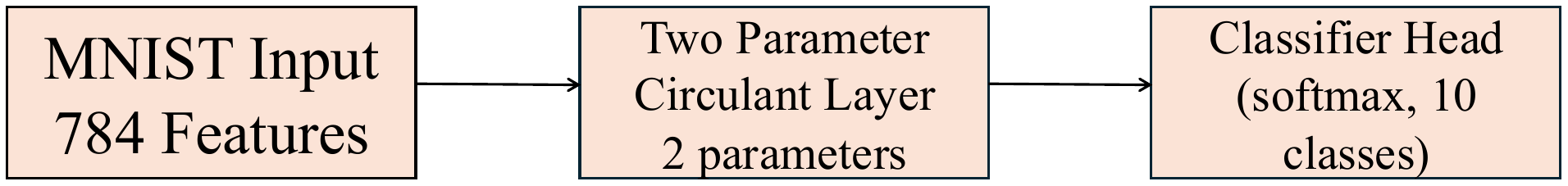}
        \caption{\textbf{Circulant Mapping (Parameter-Light)}}
        \label{fig:circulant_mapping}
    \end{subfigure}

    \caption{\textbf{Comparison of parameter usage between standard dense mapping and Two-Valued Symmetric Circulant Matrices (TVSCM).}}
    \label{fig:comparison_dense_vs_tvscm}
\end{figure}

Each layer of TVSCM performs the same basic set of operations as in traditional dense layers. The input vector is subjected to a linear transformation and then to the nonlinear activation function. The difference is in the formulation of the weight matrix, which, instead of learning each weight individually, learns it in a circulant, symmetric fashion using just two scalars. Thus, instead of learning many weights, the TVSCM learns the entire weight matrix from just two scalars. This improvement is illustrated in Fig.~\ref{fig:comparison_dense_vs_tvscm}, which compares the standard dense mapping with the two-parameter circulant mapping.

The two-valued constraint also serves as a natural form of regularization. For example, controlling the number of trainable parameters can prevent overfitting as well as facilitate the smooth convergence of the learning process. The parameter reduction helps optimize the learning process in system learning, especially when there is noise in the training set.


\begin{algorithm}[htbp]
\caption{TVSCM Forward Propagation}
\label{alg:tvscm_forward}
\begin{algorithmic}[1]

\State \textbf{Inputs:} $\mathbf{x}, a, b, n$
\State \textbf{Output:} $\mathbf{y}$

\State $\mathbf{v} \gets [a, b, a, b, \ldots]$
\State $\mathbf{v}$ has length $n$

\State $W(1,:) \gets \mathbf{v}$

\For{\mbox{$i=2$ to $n$}}
    \State $W(i,:) \gets \mathrm{shift}(W(i-1,:),1)$
\EndFor

\State $W \gets (W + W^{T})/2$
\State $\mathbf{z} \gets W\mathbf{x} + \mathbf{b}$
\State $\mathbf{y} \gets f(\mathbf{z})$

\end{algorithmic}
\end{algorithm}

This configuration ensures that each layer is learnable with just two variables, due to the application of a single transformation over all nodes in the neural network. Thus, this method is useful in improving the stability of the learning process while eliminating redundancy in the variables. Although the algorithm outlines the computational process of the TVSCM layer, its usefulness can be more appreciated when considering its deployment scenario. The following subsection will enlighten on how TVSCM can be effectively integrated into the IoMT architecture.

\section{Experimental Results}
\label{SEC:Results}

This section shows the experimental results of the proposed Two-Valued Symmetric Circulant Matrix (TVSCM) based neural network design in comparison with a conventional dense model. These experiments were conducted on two datasets: MNIST, a classification problem involving images, and MIT-BIH Arrhythmia, a classification problem involving heartbeats in the form of ECG signals, to test the model's performance and the effectiveness of the parameters. All the machine learning models in this report were built in PyTorch. 

\subsection{Evaluation on MNIST Dataset}

The MNIST dataset is used to test the efficacy of the proposed Two-Valued Symmetric Circulant Matrix (TVSCM) model in comparison to a traditional dense neural network. The traditional dense network has 623,290 trainable parameters. It achieves a maximum accuracy of 97.67\%. The proposed TVSCM model has only 7,852 trainable parameters. It achieves an accuracy of 93.54\%. This results in approximately \textit{79x} reduction in parameters as shown below:

\[
\frac{623,290}{7,852} \approx 79x
\]

There is a small reduction in accuracy by 4.13\%. However, the compression ratio is quite large. It can be used to improve computational efficiency, memory requirements, and energy consumption.

\subsubsection{Training Time Analysis}
\begin{figure}[h!]
    \centering
    \includegraphics[width=0.85\linewidth]{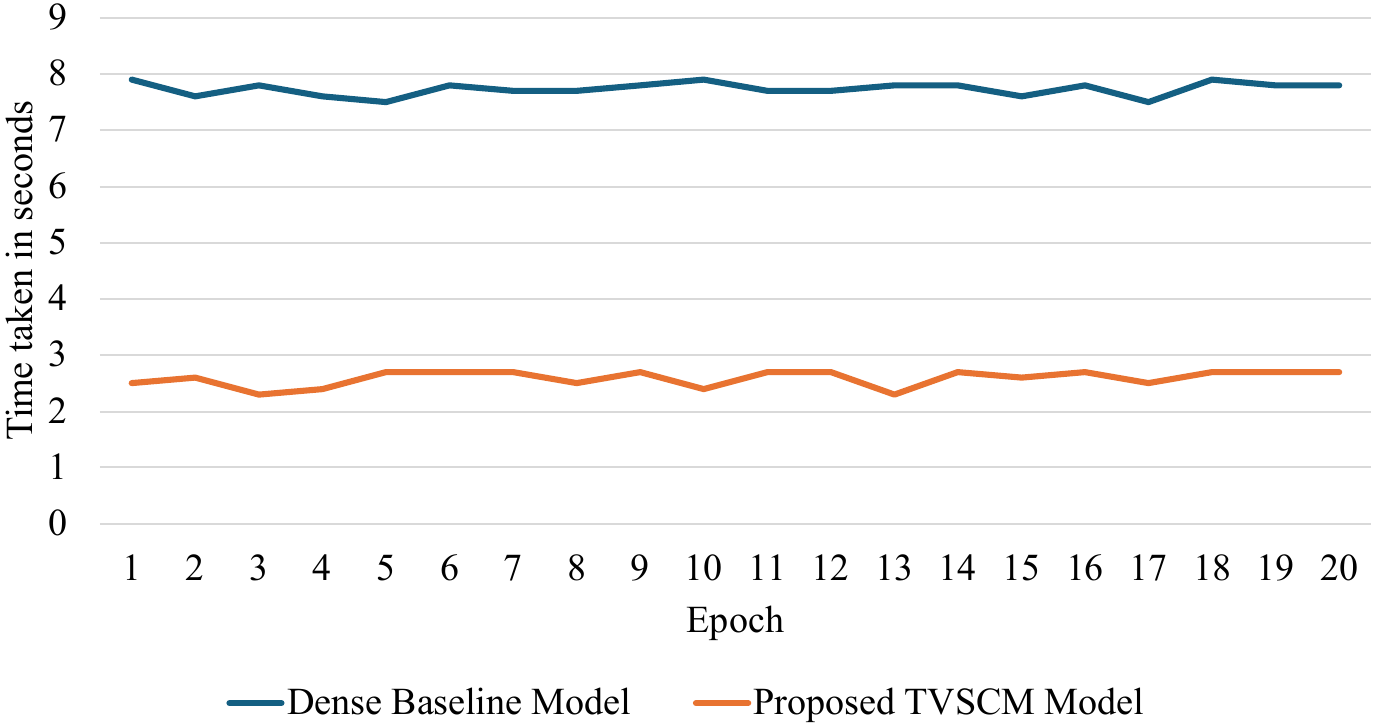}
    \caption{Time per epoch comparison between the Dense baseline and the proposed TVSCM model on the MNIST dataset using a CPU. The TVSCM model consistently achieves lower computation time across all epochs.}
    \label{fig:epoch_time_mnist}
\end{figure}

Fig.~\ref{fig:epoch_time_mnist} illustrates the training time per epoch using a CPU for the TVSCM model. The TVSCM model takes less training time than the dense network architecture for all 20 epochs. The training time of the TVSCM model is reduced because of the fewer trainable parameters.

\subsubsection{Memory Access per Inference}
To analyze the runtime behavior, memory accesses per inference were measured.

\begin{itemize}
    \item Dense model: 2.493 MB per inference
    \item TVSCM model: 0.031 MB per inference
\end{itemize}

The reduction factor is:

\[
\frac{2.493}{0.031} \approx 80.4\times
\]

\begin{figure}[h!]
    \centering
    \includegraphics[width=0.55\linewidth]{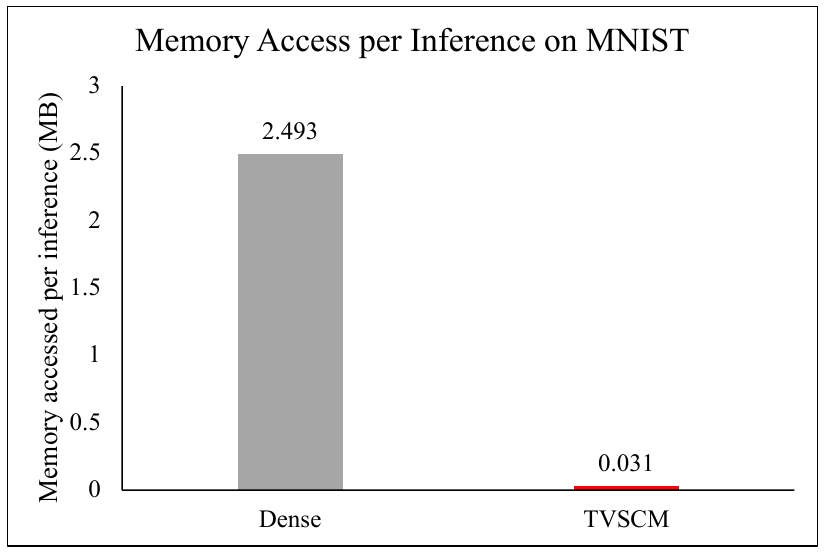}
    \caption{Memory accessed per inference on MNIST dataset(Batch Size = 32). }
    \label{fig:memory access mnist}
\end{figure}

This means the proposed model reduces memory traffic by more than $80\times$. Since memory movement is one of the primary contributors to power consumption in modern hardware, reducing memory accesses significantly improves system efficiency and makes this model more suitable for edge and embedded devices. As shown in Fig.~\ref{fig:memory access mnist}, the TVSCM model drastically reduces memory traffic compared to the dense baseline.

\subsubsection{Energy per Inference}
Another measurement was taken to calculate the energy consumption per inference to analyze its feasibility.

\begin{itemize}
    \item Dense model: 3.8 mJ per sample
    \item TVSCM model: 0.075 mJ per sample
\end{itemize}

The energy reduction factor is:

\[
\frac{3.8}{0.075} \approx 50.7\times
\]

The result shows that TVSCM model achieves 50x lower energy consumption per inference.

\begin{figure}[h!]
    \centering
    \includegraphics[width=0.55\linewidth]{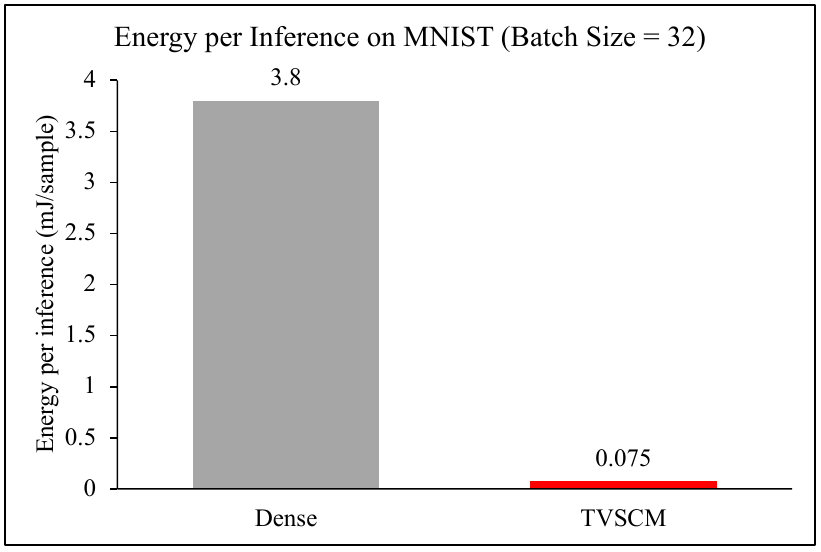}
    \caption{Energy consumption per inference on MNIST dataset(Batch Size = 32).}
    \label{fig:energy on mnist}
\end{figure}

This shows that the proposed model is suitable for battery-powered devices and IoT healthcare applications. As shown in Fig.~\ref{fig:energy on mnist}, the TVSCM model achieves nearly 50× lower energy usage.

\subsubsection{Throughput per Watt}
To assess its scalability under different conditions of workload, throughput per watt was measured under various batch sizes (1, 8, 32, and 128).

For the dense baseline model, the throughput per watt values are:

\begin{itemize}
    \item Batch 1: 212 samples/sec/W
    \item Batch 8: 260 samples/sec/W
    \item Batch 32: 265 samples/sec/W
    \item Batch 128: 258 samples/sec/W
\end{itemize}

For the proposed TVSCM model, the throughput per watt values are:

\begin{itemize}
    \item Batch 1: 1,530 samples/sec/W
    \item Batch 8: 7,500 samples/sec/W
    \item Batch 32: 13,100 samples/sec/W
    \item Batch 128: 14,250 samples/sec/W
\end{itemize}

The improvement factors at different batch sizes are approximately:

Batch 1:

\[
\frac{1530}{212} \approx 7.2\times
\]

Batch 8:

\[
\frac{7500}{260} \approx 28.8\times
\]

Batch 32:

\[
\frac{13100}{265} \approx 49.4\times
\]

Batch 128:

\[
\frac{14250}{258} \approx 55.2\times
\]

From the results, it is evident that as the batch size increases, so does the performance advantage of TVSCM.

\begin{figure}[h!]
    \centering
    \includegraphics[width=0.55\linewidth]{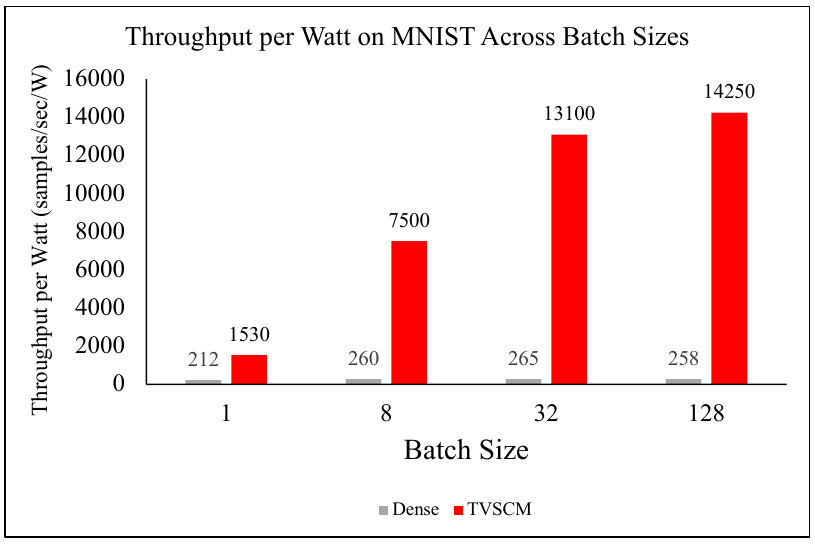}
    \caption{Throughput per watt on MNIST across batch sizes.}
    \label{fig:Throughput on mnist}
\end{figure}

In addition, the dense model reaches saturation at a batch size of 32 but does not scale up much after that. However, TVSCM shows improved scalability and efficiency as batch size increases. As shown in Fig.~\ref{fig:Throughput on mnist}, TVSCM shows improved scalability and efficiency as batch size increases.

\subsubsection{Latency Distribution Analysis}
Inference latency was also measured as part of the evaluation.

\begin{itemize}
    \item Dense model:
    \begin{itemize}
        \item Mean latency: 26.556 ms
        \item P99 latency: 26.947 ms
    \end{itemize}
    \item TVSCM model:
    \begin{itemize}
        \item Mean latency: 0.2266 ms
        \item P99 latency: 0.2544 ms
    \end{itemize}
\end{itemize}

The average speedup is:

\[
\frac{26.556}{0.2266} \approx 117\times
\]

The P99 latency improvement is approximately:

\[
\frac{26.947}{0.2544} \approx 106\times
\]

The results show that the model not only performs the inference much faster but also enjoys greater stability with a more clustered latency distribution, which is essential for real-time applications.
\begin{table*}[t]
\centering
\caption{Comprehensive Performance Comparison on MNIST}
\label{tab:mnist_comprehensive}
\renewcommand{\arraystretch}{1.2}
\setlength{\tabcolsep}{6pt}
\footnotesize
\begin{tabular}{|l|c|c|c|}
\hline
\textbf{Metric} & \textbf{Dense Model} & \textbf{TVSCM Model} & \textbf{Improvement} \\
\hhline{|=|=|=|=|}
Trainable Parameters & 623,290 & 7,852 & $\sim$79$\times$ fewer \\
\hline
Accuracy (\%) & 97.67 & 93.54 & -4.13\% \\
\hline
Memory per Inference (MB) & 2.493 & 0.031 & $\sim$80$\times$ lower \\
\hline
Energy per Inference (mJ) & 3.8 & 0.075 & $\sim$50$\times$ lower \\
\hline
Throughput per Watt (samples/sec/W) & 265 & 13,100 & $\sim$49$\times$ higher \\
\hline
Mean Latency (ms) & 26.556 & 0.2266 & $\sim$117$\times$ faster \\
\hline
P99 Latency (ms) & 26.947 & 0.2544 & $\sim$106$\times$ faster \\
\hline
\end{tabular}
\end{table*}

\subsection{Evaluation on MIT-BIH ECG Heartbeat Dataset}
To further test the proposed TVSCM model, experiments were carried out with the MIT-BIH Arrhythmia ECG dataset, which simulates a realistic scenario of how this model would be applied in a healthcare setting, where energy efficiency is critical.

The dense model has 24,709 trainable parameters, and it achieved an accuracy of 97.60\%.

The TVSCM model, on the other hand, achieved an accuracy of 93.10\% with 942 trainable parameters. This corresponds to approximately:

\[
\frac{24,709}{942} \approx 26\times
\]

reduction in model parameters. Even though there was a reduction in accuracy by 4.50\%, there were significant gains in terms of hardware efficiency.

 \subsubsection{Memory Access per Inference}
 The memory access per inference was measured to assess the efficiency during runtime.

\begin{itemize}
    \item Dense model: 0.094257 MB per inference
    \item TVSCM model: 0.003593 MB per inference
\end{itemize}

The reduction factor is:

\[
\frac{0.094257}{0.003593} \approx 26.2\times
\]

The findings show that the proposed TVSCM structure directly reduces memory traffic during inference.

\begin{figure}[h!]
    \centering
    \includegraphics[width=0.55\linewidth]{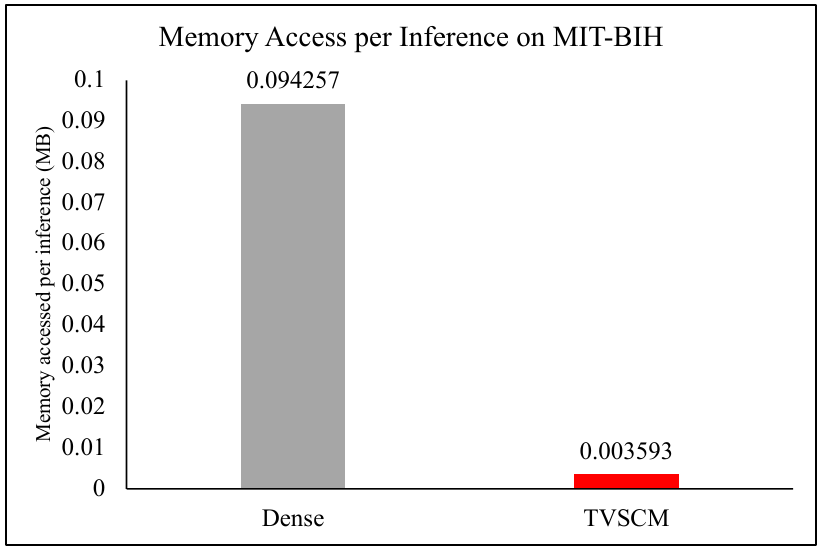}
    \caption{Memory accessed per inference on MIT-BIH ECG dataset(Batch Size = 32). }
    \label{fig:memory access mitbih}
\end{figure}

As memory access is a major contributor to system energy consumption, the proposed method improves the suitability of the system for wearable health systems. As shown in Fig.~\ref{fig:memory access mitbih}, the TVSCM reduces memory requirements by a significant amount compared to the dense model.

\subsubsection{Energy per Inference}
The energy consumption per inference was also measured.

\begin{itemize}
    \item Dense model: 2.96 mJ per sample
    \item TVSCM model: 0.0265 mJ per sample
\end{itemize}

The reduction factor is:

\[
\frac{2.96}{0.0265} \approx 112\times
\]

The findings show that the proposed TVSCM model reduces energy consumption by over 100 times during ECG classification.

\begin{figure}[h!]
    \centering
    \includegraphics[width=0.55\linewidth]{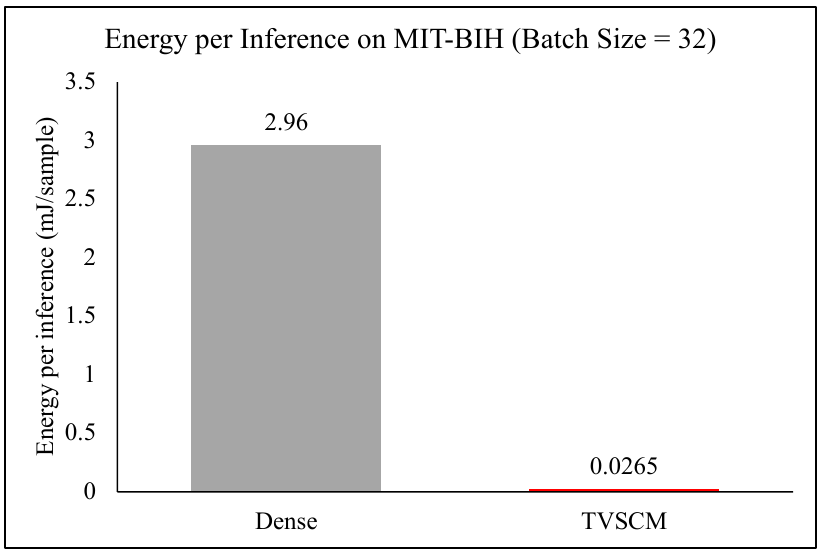}
    \caption{Energy consumption per inference on MITBIH ECG dataset(Batch Size = 32).}
    \label{fig:energy on mitbih}
\end{figure}

This is particularly significant in health systems such as cardiac monitoring devices, where energy efficiency is a major factor in the viability of the system. As shown in Fig.~\ref{fig:energy on mitbih}, the TVSCM model has shown improved computational efficiency under power-constrained conditions.

\subsubsection{Throughput per Watt}
The throughput per watt is used to assess the scalability of the model at different batch sizes: 1, 8, 32, and 128.

For the dense baseline model:

\begin{itemize}
    \item Batch 1: 1,700 samples/sec/W
    \item Batch 8: 5,100 samples/sec/W
    \item Batch 32: 6,350 samples/sec/W
    \item Batch 128: 6,300 samples/sec/W
\end{itemize}

For the TVSCM model:

\begin{itemize}
    \item Batch 1: 6,800 samples/sec/W
    \item Batch 8: 12,300 samples/sec/W
    \item Batch 32: 37,500 samples/sec/W
    \item Batch 128: 72,500 samples/sec/W
\end{itemize}

The improvement factors are approximately:

Batch 1:

\[
\frac{6800}{1700} \approx 4\times
\]

Batch 8:

\[
\frac{12300}{5100} \approx 2.4\times
\]

Batch 32:

\[
\frac{37500}{6350} \approx 5.9\times
\]

Batch 128:

\[
\frac{72500}{6300} \approx 11.5\times
\]

The results show that the advantage of TVSCM increases as the batch size increases. Although the dense model plateaus at an early stage and does not scale much beyond the 32-batch size, the structured TVSCM model continues to improve throughput efficiency. This shows that the proposed model has addressed the memory bottleneck issue and scaled better at higher workload conditions.

\begin{figure}[h!]
    \centering
    \includegraphics[width=0.55\linewidth]{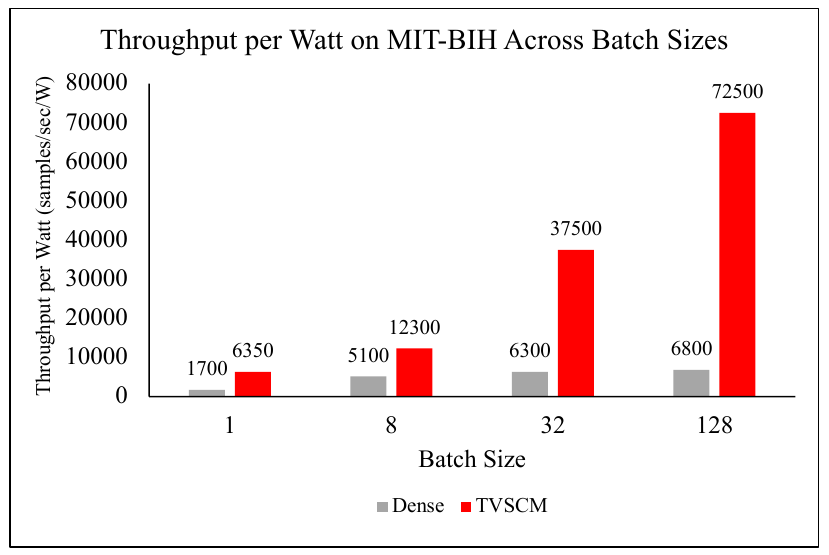}
    \caption{Throughput per watt on MIT-BIH across batch sizes.}
    \label{fig:Throughput on mitbih}
\end{figure}

The TVSCM model shows better scalability and increasing efficiency as the batch size increases. Fig.~\ref{fig:Throughput on mitbih} shows TVSCM scales efficiently with increasing batch size, unlike the dense baseline.

\subsubsection{Latency Distribution Analysis}
To evaluate the performance in the context of real-time inference, inference latency was studied for the performance evaluation of the proposed model, TVSCM, using the MIT-BIH ECG dataset.

For the dense baseline model:

\begin{itemize}
    \item Mean latency: 15.2171 ms
    \item P50 latency: 15.2158 ms
    \item P90 latency: 15.2189 ms
    \item P99 latency: 15.2420 ms
\end{itemize}

For the proposed TVSCM model:

\begin{itemize}
    \item Mean latency: 0.1664 ms
    \item P50 latency: 0.1660 ms
    \item P90 latency: 0.1681 ms
    \item P99 latency: 0.1823 ms
\end{itemize}

The average speedup is:

\[
\frac{15.2171}{0.1664} \approx 91.5\times
\]

The P99 latency improvement is:

\[
\frac{15.2420}{0.1823} \approx 83.6\times
\]

From the results, it is clear that the proposed TVSCM model is able to achieve around $92\times$ faster average inference latency and over $83\times$ improvement in worst-case latency. The distribution of the latency for P50, P90, and P99 is quite close, which is a good property for a real-time system since any delay in the biomedical system may impact the safety of the patients. TVSCM is able to achieve much lower mean and worst-case latency compared to the dense baseline.

\begin{table*}[t]
\centering
\caption{Comprehensive Performance Comparison on MIT-BIH ECG}
\label{tab:mitbih_comprehensive}
\renewcommand{\arraystretch}{1.2}
\setlength{\tabcolsep}{6pt}
\footnotesize
\begin{tabular}{|l|c|c|c|}
\hline
\textbf{Metric} & \textbf{Dense Model} & \textbf{TVSCM Model} & \textbf{Improvement} \\
\hhline{|=|=|=|=|}
Trainable Parameters & 24,709 & 942 & $\sim$26$\times$ fewer \\
\hline
Accuracy (\%) & 97.60 & 93.10 & -4.50\% \\
\hline
Memory per Inference (MB) & 0.094257 & 0.003593 & $\sim$26$\times$ lower \\
\hline
Energy per Inference (mJ) & 2.96 & 0.0265 & $\sim$112$\times$ lower \\
\hline
Throughput per Watt (Batch 32) & 6,350 & 37,500 & $\sim$5.9$\times$ higher \\
\hline
Mean Latency (ms) & 15.2171 & 0.1664 & $\sim$91.5$\times$ faster \\
\hline
P99 Latency (ms) & 15.2420 & 0.1823 & $\sim$83.6$\times$ faster \\
\hline
\end{tabular}
\end{table*}

\subsection{Cross-Dataset System-Level Discussion}
The results obtained in both MNIST and MIT-BIH experiments show an important and consistent pattern on the system level. Despite the fact that the two datasets are related to different application domains, namely image classification and biomedical signal analysis, the performance of the proposed TVSCM model is consistent and predictable in both scenarios.

\begin{figure}[h!]
    \centering
    \includegraphics[width=0.7\textwidth]{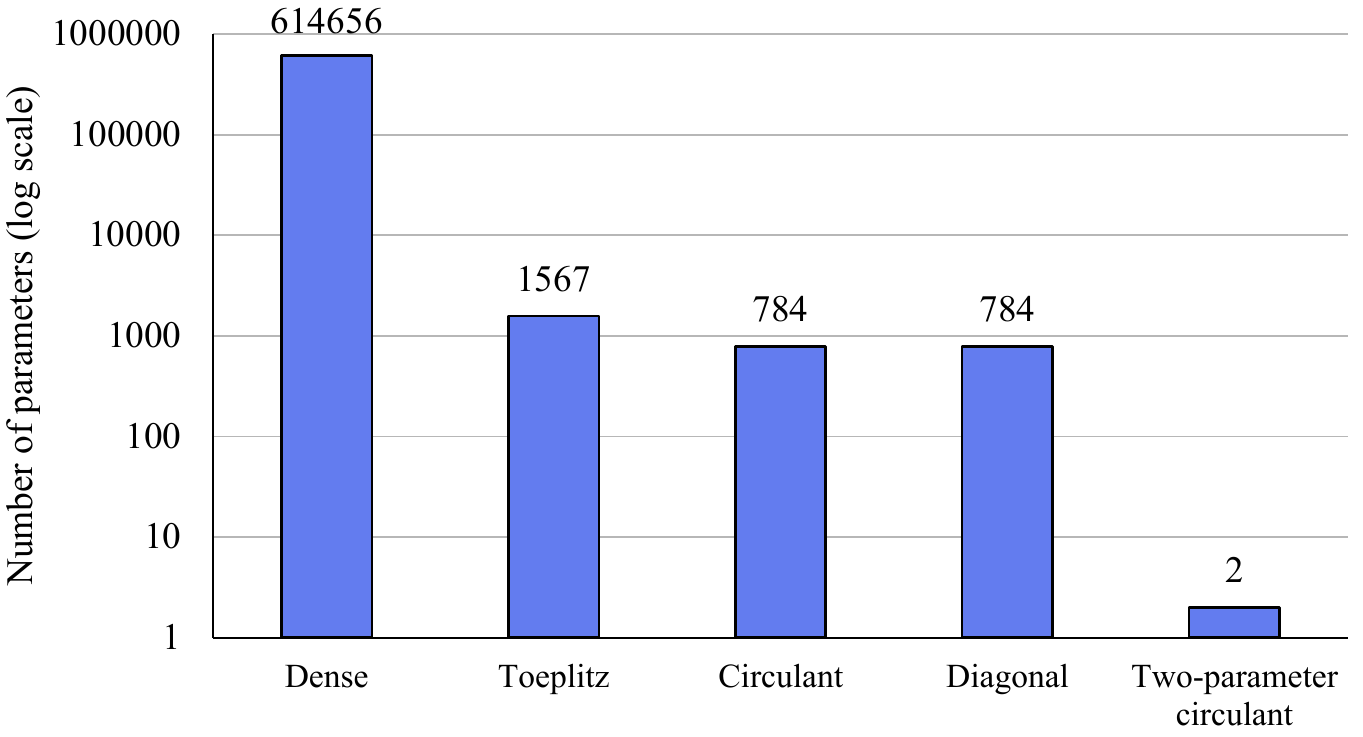}
    \caption{Parameter comparison across matrix structures for n=784.}
    \label{Experimental_Results_Bar_Graph}
\end{figure}

As indicated by Tables 7, and as shown by the log-scale parameter comparison in Fig.~\ref{Experimental_Results_Bar_Graph}, the structured matrix formulation reduces model size substantially for both datasets. This work, apart from being a theoretical achievement, also represents an improvement in terms of system performance. By reducing the size of the model, it is possible to decrease the amount of data that is required to be stored, accessed, and transported during execution, thus exerting an immediate influence on the interaction with hardware. 

The experiments carried out show that there is an improvement in various parameters of the system due to reduced memory movements. It has been seen that reduced memory movements result in reduced power consumption, as memory access is considered to be one of the most power-consuming activities in modern processors. This improvement does not require any hardware modifications, as seen in the experiments.

Moreover, the latency characteristics further confirm the benefits that are achieved at the system level. The model that is proposed has demonstrated its capability to decrease average running times as well as improve stability in performance. Furthermore, the decrease in latency has improved predictability in the inference process, which is critical in real-time systems, as any delay in response may compromise the reliability of intelligent systems, such as those used in healthcare monitoring.

An evaluation of throughput, based on varying batch sizes, has also demonstrated further benefits that are achieved by using the proposed model. This is because, while the dense model shows plateauing performance as batch size is increased, the structured TVSCM model shows improved efficiency in scalability. This further confirms that memory-bound constraints are being addressed by the proposed model, allowing for improved efficiency in computation while keeping power consumption constant.

An interesting observation can be noted with respect to the proposed methodology when it is evaluated. It can be noted that the benefits remain consistent across various datasets. This proves that the benefits provided by the proposed structured matrix approach are generic, thus making it more practical from an applicability point of view with respect to various AI deployment scenarios. Moreover, it can be evaluated that with various datasets, it can be noted that the proposed TVSCM does not only provide an efficient solution from a hardware point of view, but it can also provide an efficient solution with respect to parameter reduction.

\begin{table}[h!]
\centering
\caption{Characterization of Dense vs. TVSCM Models}
\begin{tabular}{|l|c|c|}
\hline
\textbf{Aspect} & \textbf{Dense Model} & \textbf{TVSCM Model} \\
\hhline{|=|=|=|}
Matrix Type & Fully connected dense & Two-valued symmetric circulant \\
\hline
MNIST Parameters & 623,290 & 7,852 \\
\hline
ECG Parameters & 24,709 & 942 \\
\hline
Compression Ratio & None & 26--79$\times$ fewer parameters \\
\hline
Unique Values per Matrix & Many & Only two values \\
\hline
Symmetry & No symmetry & Symmetric and circulant \\
\hline
Computation Cost & High (O($n^2$)) & Lower (O($n \log n$)) \\
\hline
Memory Usage & High & Very low \\
\hline
Training Speed & Normal & Faster \\
\hline
Inference Use & Cloud/desktop & Edge/wearable devices \\
\hline
Datasets Tested & MNIST, ECG & MNIST, ECG \\
\hline
\end{tabular}
\end{table}

\section{Conclusion and future research}
\label{SEC:Conclusion}
The experiment makes it clear how powerful Two-Valued Symmetric Circulant Matrices (TVSCM) can be as a lightweight solution for effectively compressing fully connected layers in deep learning models. The symmetry pattern of two values per fully connected layer in circulants makes it very efficient, with reduced complexity and storage requirements. Though this experiment yields very promising results for effectively using Two-Valued Symmetric Circulant Matrices as just another architectural solution for compressing models for edge and Internet-of-Medical-Things devices, several areas remain to be explored. Extending these to other models, such as convolutional neural networks and recurrent neural networks/transformers, would be quite interesting. More organized approaches towards adaptive/multi-level compressions, along with their noise/attention/real-time data robustness research arenas, would be very interesting to pursue. Hardware-level joint development to improve suitability for edge devices would be quite valuable.
\bibliographystyle{unsrt}  
\bibliography{references}  

\section*{Authors' Biographies}

\noindent
\begin{wrapfigure}{l}{0.22\textwidth}
    \vspace{-6pt}
    \includegraphics[width=0.22\textwidth]{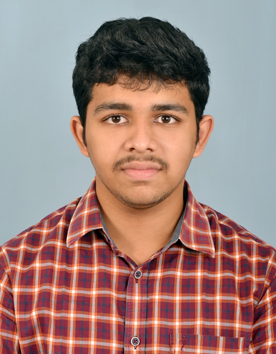}
    \vspace{-10pt}
\end{wrapfigure}
\noindent \textbf{Jayakrishna Amathi} received the Bachelor of Technology degree in Computer Science and Engineering from R.V.R.\& J.C.\ College of Engineering, Guntur, India, in 2023, and the Master of Science degree in Computer Science from the University of North Texas, Denton, TX, in 2025. He is currently pursuing the Ph.D.\ degree in Computer Science at the University of North Texas, Denton, TX, under the supervision of Dr.\ Saraju P.\ Mohanty, with a GPA of 4.00. His research interests include machine learning, deep learning, structured matrix models, and AI-driven systems for healthcare and edge intelligence applications. He has experience in building and evaluating neural network models, data pipelines, and experimentation workflows using Python, PyTorch, and TensorFlow. He has explored Generative AI concepts including large language models (LLMs), prompt engineering, and retrieval-augmented generation (RAG)-based systems. He is currently serving as a Teaching Assistant in the Department of Computer Science and Engineering at the University of North Texas. He is an author of 3 peer-reviewed journal and conference publications.

\bigskip

\noindent
\begin{wrapfigure}{l}{0.22\textwidth}
    \vspace{-6pt}
    \includegraphics[width=0.22\textwidth]{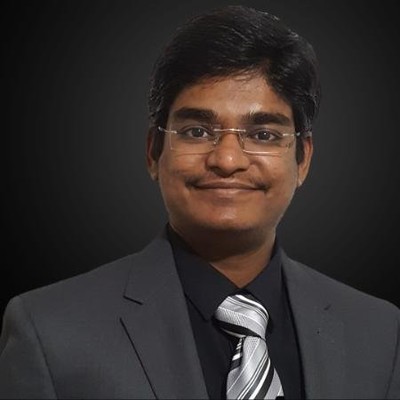}
    \vspace{-10pt}
\end{wrapfigure}
\noindent \textbf{Venkata P. Yanambaka} (M'19) received the Bachelor of Technology degree in Computer Science and Engineering from Priyadarshini College of Engineering and Technology, Nellore, India, the Master of Science degree in Computer Engineering from the University of North Texas, Denton, TX, and the Ph.D.\ degree in Computer Science and Engineering from the University of North Texas, Denton, TX, in 2019, under the supervision of Dr.\ Saraju P.\ Mohanty. His dissertation, titled ``Exploring Physical Unclonable Functions for Efficient Hardware Assisted Security in IoT,'' was awarded the Best Poster Award at IEEE MetroCon 2017. He is currently an Assistant Professor of Computer Science at Texas Woman's University, Denton, TX. His research interests are in the Internet of Things (IoT), hardware-assisted security primitives, Physical Unclonable Functions (PUFs), and security in IoT architectures. He has authored over 35 peer-reviewed publications, including multiple journal and transaction articles and a book chapter. He has served as a reviewer for several peer-reviewed journals and conferences. He has been an active member of IEEE since 2012.

\bigskip

\noindent
\begin{wrapfigure}{l}{0.22\textwidth}
    \vspace{-6pt}
    \includegraphics[width=0.22\textwidth]{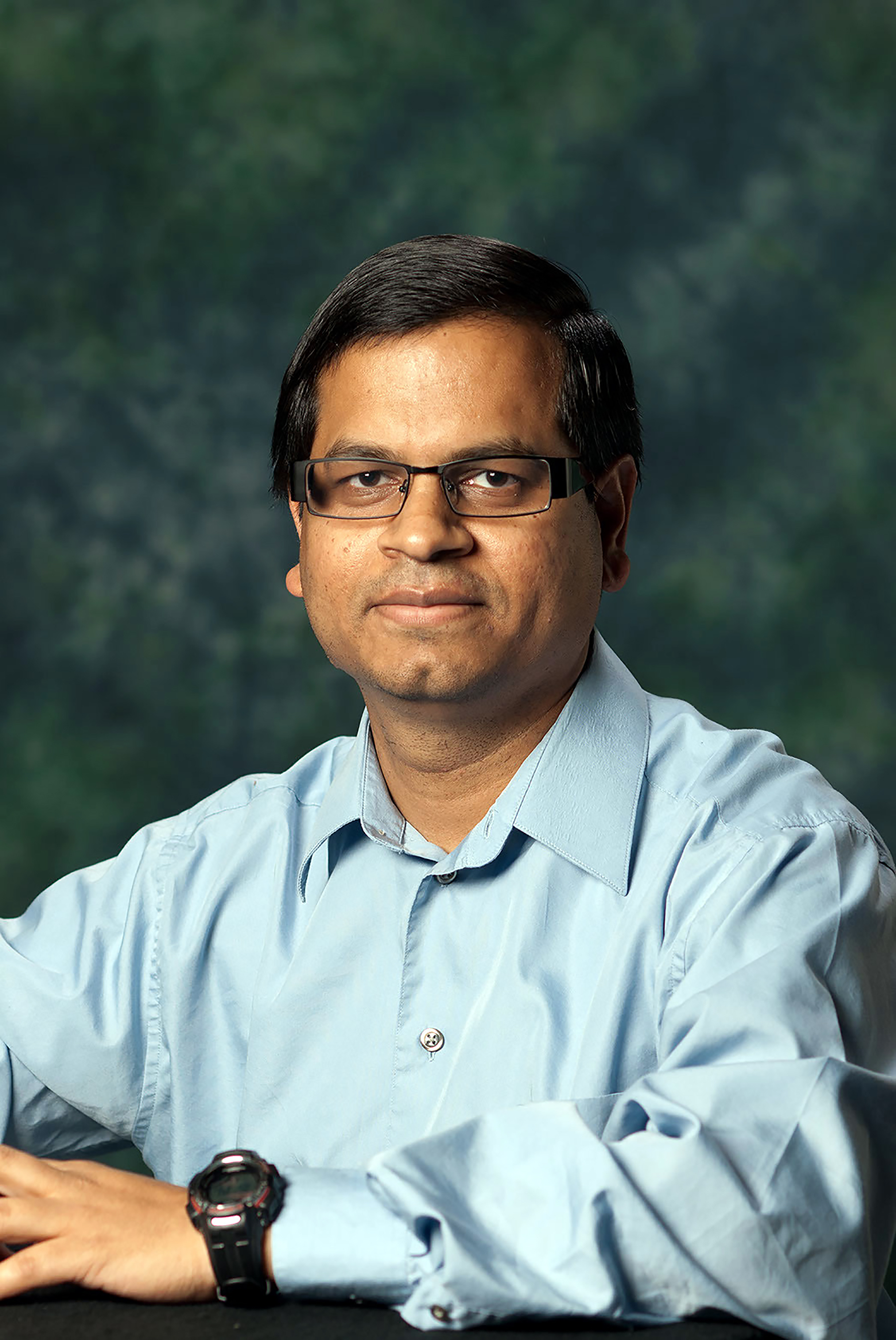}
    \vspace{-10pt}
\end{wrapfigure}
\noindent \textbf{Saraju P. Mohanty} received the bachelor's degree (Honors) in electrical engineering from the Orissa University of Agriculture and Technology, Bhubaneswar, in 1995, the master's degree in Systems Science and Automation from the Indian Institute of Science, Bengaluru, in 1999, and the Ph.D. degree in Computer Science and Engineering from the University of South Florida, Tampa, in 2003. He is a Professor with the University of North Texas. His research is in ``Smart Electronic Systems'' which has been funded by National Science Foundations (NSF), Semiconductor Research Corporation (SRC), U.S. Air Force, NIDILRR, IUSSTF, and Mission Innovation. He has authored 550 research articles, 5 books, and 10 granted and pending patents. His Google Scholar h-index is 62 and i10-index is 300 with 17,000 citations. He is regarded as a visionary researcher on Smart Cities technology in which his research deals with security and energy aware, and AI/ML-integrated smart components. He introduced the Secure Digital Camera (SDC) in 2004 with built-in security features designed using Hardware Assisted Security (HAS) or Security by Design (SbD) principle. He is widely credited as the designer for the first digital watermarking chip in 2004 and first the low-power digital watermarking chip in 2006. He is a recipient of 21 best paper awards, Fulbright Specialist Award in 2021, IEEE Consumer Electronics Society Outstanding Service Award in 2020, the IEEE-CS-TCVLSI Distinguished Leadership Award in 2018, and the PROSE Award for Best Textbook in Physical Sciences and Mathematics category in 2016. He has delivered 31 keynotes and served on 15 panels at various International Conferences. He has been serving on the editorial board of several peer-reviewed international transactions/journals, including IEEE Transactions on Big Data (TBD), IEEE Transactions on Computer-Aided Design of Integrated Circuits and Systems (TCAD), IEEE Transactions on Consumer Electronics (TCE), and ACM Journal on Emerging Technologies in Computing Systems (JETC). He has been the Editor-in-Chief (EiC) of the IEEE Consumer Electronics Magazine (MCE) during 2016--2021. He served as the Chair of Technical Committee on Very Large Scale Integration (TCVLSI), IEEE Computer Society (IEEE-CS) during 2014--2018 and on the Board of Governors of the IEEE Consumer Electronics Society during 2019--2021. He serves on the steering, organizing, and program committees of several international conferences. He is the steering committee chair/vice-chair for the IEEE International Symposium on Smart Electronic Systems (IEEE-iSES), the IEEE-CS Symposium on VLSI (ISVLSI), and the OITS International Conference on Information Technology (OCIT). He has supervised 3 post-doctoral researchers, 19 Ph.D. dissertations, 29 M.S. theses, and 41 undergraduate projects.

\bigskip

\noindent
\begin{wrapfigure}{l}{0.22\textwidth}
    \vspace{-6pt}
    \includegraphics[width=0.22\textwidth]{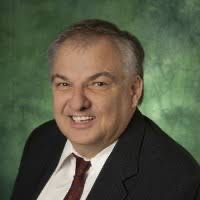}
    \vspace{-10pt}
\end{wrapfigure}
\noindent \textbf{Elias Kougianos} received a BSEE from the University of Patras, Greece in 1985 and an MSEE in 1987, an MS in Physics in 1988 and a Ph.D.\ in EE in 1997, all from Louisiana State University. From 1988 through 1998 he was with Texas Instruments, Inc., in Houston and Dallas, TX. In 1998 he joined Avant! Corp.\ (now Synopsys) in Phoenix, AZ as a Senior Applications engineer and in 2000 he joined Cadence Design Systems, Inc., in Dallas, TX as a Senior Architect in Analog/Mixed-Signal Custom IC design. He has been at UNT since 2004. He is a Professor in the Department of Electrical Engineering, at the University of North Texas (UNT), Denton, TX. His research interests are in the area of Analog/Mixed-Signal/RF IC design and simulation and in the development of VLSI architectures for multimedia applications. He is an author of over 200 peer-reviewed journal and conference publications.

\end{document}